\documentclass[10pt,twocolumn,letterpaper]{article}

\usepackage{iccv}
\usepackage{times}
\usepackage{epsfig}
\usepackage{graphicx}
\usepackage{amsmath}
\usepackage{amssymb}
\usepackage{caption}
\usepackage{subcaption}
\usepackage{array}

\usepackage{booktabs}

\graphicspath{{../figures/}}

\usepackage[pagebackref=true,breaklinks=true,letterpaper=true,colorlinks,bookmarks=false]{hyperref}

\iccvfinalcopy 

 \date{}

\ificcvfinal\fi
\begin{document}

\title{SCNet: Learning Semantic Correspondence}

\author{Kai Han\textsuperscript{1} \quad\quad Rafael S. Rezende\textsuperscript{4,5} \quad\quad Bumsub Ham\textsuperscript{2} \quad\quad Kwan-Yee K. Wong\textsuperscript{1} \vspace*{0.1cm}\\
Minsu Cho\textsuperscript{3} \quad\quad\quad Cordelia Schmid\textsuperscript{4,}\thanks{Univ. Grenoble Alpes, Inria, CNRS, Grenoble INP, LJK, 38000 Grenoble, France.} \quad\quad\quad Jean Ponce\textsuperscript{5,4} \vspace*{0.2cm}\\
\textsuperscript{1}The University of Hong Kong \quad\quad \textsuperscript{2}Yonsei Univ. \quad\quad \textsuperscript{3}POSTECH \quad\quad \textsuperscript{4}Inria \\ \textsuperscript{5}Department of Computer Science, ENS / CNRS / PSL Research University }

\maketitle

\begin{abstract} This paper addresses the problem of establishing {\em semantic correspondences} between images depicting different instances of the same object or scene category.  Previous approaches focus on either combining a spatial regularizer with hand-crafted features, or learning a correspondence model for appearance only. We propose instead a convolutional neural network architecture, called {\em SCNet}, for learning a {\em geometrically} plausible model for semantic correspondence.  SCNet uses region proposals as matching primitives, and explicitly incorporates geometric consistency in its loss function.  It is trained on image pairs obtained from the PASCAL VOC 2007 keypoint dataset, and a comparative evaluation on several standard benchmarks demonstrates that the proposed approach substantially outperforms both recent deep learning architectures and previous methods based on hand-crafted features.  \end{abstract}

\section{Introduction}
Our goal in this paper is to establish {\em semantic correspondences} across images that contain different instances of the same object or scene category, and thus feature much larger changes in appearance and spatial layout than the pictures of the {\em same} scene used in {\em stereo vision}, which we take here to include broadly not only classical (narrow-baseline) stereo fusion (e.g., ~\cite{okutomi1993multiple,rhemann2011fast}), but also optical flow computation (e.g.,~\cite{horn1993determining,weinzaepfel2015deepmatching,weinzaepfel2013deepflow}) and wide-baseline matching (e.g.,~\cite{matas2004robust,yang2014daisy}). Due to such a large degree of variations, the problem of semantic correspondence remains very challenging. 
Most previous approaches to semantic correspondence~\cite{BristowVL15,hur2015generalized,kim2013deformable,liu2011sift,taniai2016joint,yang2014daisy} focus on combining an effective spatial regularizer with hand-crafted features such as SIFT~\cite{lowe2004distinctive}, DAISY~\cite{tola2010daisy} or HOG~\cite{dalal2005histograms}.  With the remarkable success of deep learning approaches in visual recognition, several learning-based methods have also been proposed for both stereo vision~\cite{fischer2015flownet,han2015matchnet,vzbontar2014computing,zbontar2016stereo} and semantic correspondence~\cite{UCN16,kim2017fcss,zhou2016learning}. Yet, none of these methods exploits the geometric consistency constraints that have proven to be a key factor to the success of their hand-crafted counterparts. Geometric regularization, if any, occurs during post-processing but not during learning (e.g., ~\cite{vzbontar2014computing,zbontar2016stereo}).

\begin{figure}[t]
\centering
\includegraphics[width=0.95\linewidth]{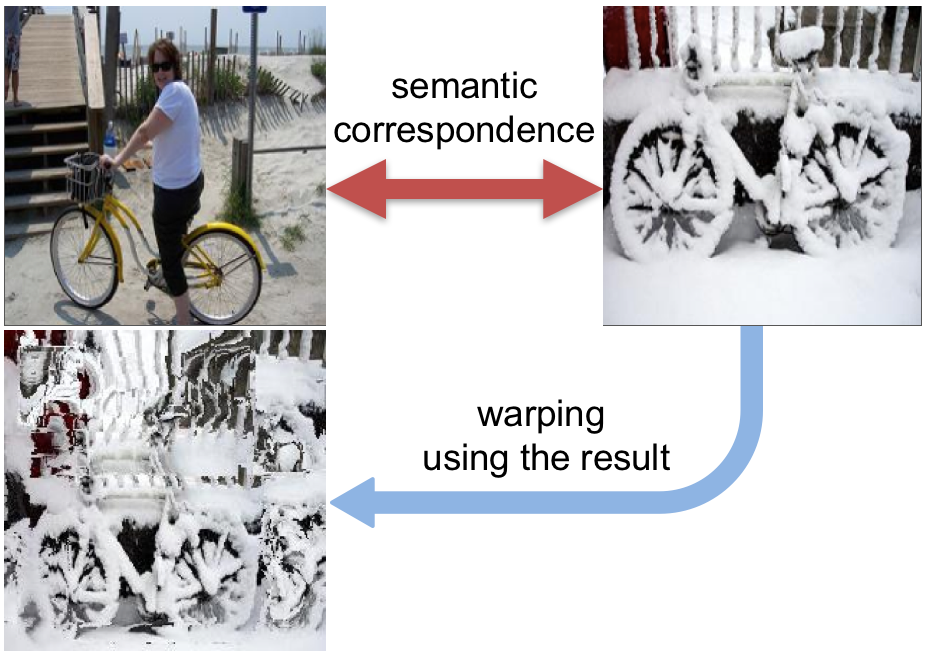}
\caption{Learning semantic correspondence. We propose a convolutional neural network, SCNet, to learn semantic correspondence using both appearance and geometry. This allows us to handle a large degree of intra-class and scene variations. This figure shows a pair of input images (top) and a warped image (bottom) using its semantic correspondence by our method. (Best viewed in color.)}
\label{fig:teaser}
\end{figure}

In this paper we propose a convolutional neural network (CNN) architecture, called {\em SCNet}, for learning geometrically plausible semantic correspondence (Figure~\ref{fig:teaser}).  Following the {\em proposal flow} approach to semantic correspondence of Ham \etal\cite{HCSP16}, we use object proposals~\cite{manen2013prime,uijlings2013selective,zitnick2014edge} as matching primitives, and explicitly incorporate the geometric consistency of these proposals in our loss function.  Unlike~\cite{HCSP16} with its hand-crafted features, however, we train our system in an end-to-end manner using image pairs extracted from the PASCAL VOC 2007 keypoint dataset~\cite{everingham2008pascal}.  A comparative evaluation on several standard benchmarks demonstrates that the proposed approach substantially outperforms both recent deep architectures and previous methods based on hand-crafted features.

Our main contributions can be summarized as follows:\vspace{-0.1cm}
\begin{itemize}
\item We introduce a simple and efficient model for learning to match regions using both appearance and geometry.\vspace{-0.1cm}
\item We propose a convolutional neural network, SCNet, to learn semantic correspondence with region proposals.\vspace{-0.1cm}
\item We achieve state-of-the-art results on several benchmarks, clearly demonstrating the advantage of  learning both appearance and geometric terms.
\end{itemize}


\section{Related work}
Here we briefly describe representative approaches related to semantic correspondence. 

\vspace{-0.3cm}
\paragraph{Semantic correspondence.}
SIFT Flow~\cite{liu2011sift} extends classical optical flow to establish  correspondences across  similar but different scenes.~It uses dense SIFT descriptors to capture semantic information beyond naive color values, and leverages a hierarchical optimization technique in a coarse-to-fine pipeline for efficiency. Kim~\emph{et al}.~\cite{kim2013deformable} and Hur~\emph{et al}.~\cite{hur2015generalized} propose more efficient generalizations of SIFT Flow. Instead of using SIFT features, Yang~\emph{et al}.~\cite{yang2014daisy} use DAISY~\cite{tola2010daisy} for an efficient descriptor extraction. Inspired by an exemplar-LDA approach~\cite{hariharan2012discriminative}, Bristow~\emph{et al}.~\cite{BristowVL15} use whitened SIFT descriptors, making semantic correspondence robust to background clutter. Recently, Ham~\emph{et al}.~\cite{HCSP16} introduces proposal flow that uses object proposals as matching elements for semantic correspondence robust to scale and clutter. This work shows that the HOG descriptor gives better matching performance than deep learning features~\cite{krizhevsky2012imagenet, simo2015discriminative}. Taniai~\emph{et al}.~\cite{taniai2016joint} also use HOG descriptors, and show that jointly performing cosegmentation and establishing dense correspondence are helpful in both tasks. Despite differences in feature descriptors and optimization schemes, these semantic correspondence approaches use a spatial regularizer to ensure flow smoothness on top of hand-crafted or pre-trained features.

\vspace{-0.3cm}
\paragraph{Deep learning for correspondence.}
Recently, CNNs have been applied to classical dense correspondence problems such as optical flow and stereo matching to learn feature descriptors~\cite{Zagoruyko15CVPR, vzbontar2014computing, zbontar2016stereo} or similarity functions~\cite{han2015matchnet, Zagoruyko15CVPR, vzbontar2014computing}. FlowNet~\cite{fischer2015flownet} uses an end-to-end scheme to learn optical flow with a synthetic dataset, 
and several recent approaches also use supervision from reconstructed 3D scenes and stereo pairs~\cite{han2015matchnet, Zagoruyko15CVPR, vzbontar2014computing, zbontar2016stereo}. MC-CNN~\cite{vzbontar2014computing} and its efficient extension~\cite{zbontar2016stereo} train CNN models to predict how well two image patches match and use this information to compute the stereo matching cost. DeepCompare~\cite{Zagoruyko15CVPR} learns a similarity function for patches directly from images of a 3D scene, which allows for various types of geometric and photometric transformations~(e.g., rotation and illumination changes). These approaches are inherently limited to matching images of the same physical object/scene. In contrast, Long~\emph{et al}.~\cite{long2014convnets} use CNN features pre-trained for ImageNet classification tasks (due to a lack of available datasets for learning semantic correspondence) with performance comparable to SIFT flow. To overcome the difficulty in obtaining ground truth for semantic correspondence, Zhou~\emph{et al}.~\cite{zhou2016learning} leverage 3D models, and uses flow consistency between 3D models and 2D images as a supervisory signal to train a CNN. Another approach to generating ground truth is to directly augment the data by densifying sparse keypoint annotations using warping~\cite{HCSP16, kanazawa2016warpnet}. 
The universal correspondence network~(UCN) of Choy {\em et al.}~\cite{UCN16} learns semantic correspondence using an architecture similar to~\cite{zbontar2016stereo}, but adds a convolutional spatial transformer networks for improved robustness to rotation and scale changes. Kim~\emph{et al}.~\cite{kim2017fcss} introduce a convolutional descriptor using self-similarity, called fully convolutional self-similarity
(FCSS), and combine the learned semantic descriptors with the proposal flow~\cite{HCSP16} framework. These approaches to learning semantic correspondence~\cite{UCN16, zhou2016learning} or semantic descriptors~\cite{kim2017fcss} typically perform better than traditional hand-crafted ones. Unlike our method, however, they do not incorporate geometric consistency between regions or object parts in the learning process. 



\section{Our approach}
We consider the problem of learning to match regions with arbitrary positions and sizes in pairs of images. This setting is general enough to cover all cases of region sampling used in semantic correspondence: sampling a dense set of regular local regions as in typical dense correspondence~\cite{BristowVL15,kim2013deformable,liu2011sift,tau2014dense} as well as employing multi-scale object proposals~\cite{arbelaez2014multiscale,hosang2015what,manen2013prime,uijlings2013selective,zitnick2014edge}.
In this work, following proposal flow~\cite{HCSP16}, 
we focus on establishing correspondences between object proposal boxes.  

\subsection{Model}
Our basic model for matching starts from the probabilistic Hough matching (PHM) approach of~\cite{CKSP15,HCSP16}. In a nutshell, given some potential match $m$ between
two regions, and the supporting data $D$ (a set of  
potential matches), the PHM model can be written as
\begin{align}
P(m|D) &= \sum_x P(m|x, D) P(x|D)  \nonumber \\
		&= P_a(m)\sum_x P_g(m|x)P(x|D),	
\end{align}
where $x$ is the offset (\eg, position and scale change) between all potential matches $m=[r,s]$ of two regions $r$ and $s$. $P_a(m)$ is the probability that the match between two regions is correct based on
appearance only, and $P_g(m|x)$ is the probability based
on geometry only, computed using the offset $x$\footnote{We suppose that appearance matching is independent of geometry matching and the offset.}.
PHM computes a matching score by replacing geometry prior $P(x|D)$ with the Hough voting $h(x|D)$~\cite{CKSP15}:  
\begin{equation}
h(x|D) = \sum_{m'\in D} P_a(m')P_g(m'|x).
\end{equation}
This turns out to be an effective spatial matching model that combines appearance similarity with global geometric consistency measured by letting all matches vote on the potential offset $x$~\cite{CKSP15,HCSP16}. 

In our learning framework, we consider similarities rather than
probabilities, and rewrite the PHM score for the match $m$ as
\begin{equation} 
\begin{array}{lcl}
z(m,w)\!\!\!\!&=\displaystyle f(m,w)\sum_x g(m,x)\sum_{m'\in D} f(m',w)g(m',x)\\
&=\displaystyle f(m,w) \sum_{m'\in D} [\sum_x g(m,x) g(m',x)] f(m',w),
\end{array}
\label{eq:eq1}
\end{equation} 
where $f(m,w)$ is a parameterized appearance similarity function between
the two regions in the potential match $m$, $x$ is as before an offset variable
(position plus scale), and $g(m,x)$ measures the geometric compatibility   
between the match $m$ and the offset $x$.

Now assuming that we have a total number of $n$ potential matches, and
identifying matches with their indices, we can
rewrite this score as
\begin{eqnarray}
  z(m,w)=f(m,w)\sum_{m'} K_{mm'} f(m',w),\nonumber\\ 
  \text{where}\,\, 
  K_{mm'}= \sum_x g(m,x) g(m',x),\label{eq:s_func}\end{eqnarray}
and the $n\times n$ matrix $K$ is the {\em kernel matrix}
associated with the feature vector $\varphi(m)=[g(m,x_1),
\ldots,g(m,x_s)]^T$, where $x_1$ to $x_s$ form the finite 
set of values that the offset variable $x$ runs
over: indeed $K_{mm'}=\varphi(m)\cdot\varphi(m')$.\footnote{Putting it all together in an $n$-vector of scores,
  this can also be rewritten as
  $z(w)=f(w)\odot Kf(w)$,
  where $z(w)=(z(1,w),\ldots,z(n,w))^T$,
  ``$\odot$'' stands for the elementwise product between vectors,
  and $f(w)=(f(1,w),\ldots,f(n,w))^T$.}

Given training pairs of images with associated true and false matches, 
we can learn our similarity function by minimizing with respect to $w$
\begin{equation}
E(w)= \sum_{m=1}^n l[y_m,z(m,w)]
+\lambda\Omega(w), 
\label{eq:main}
\end{equation} 
where $l$ is a loss function, $y_m$ is the the
ground-truth label (either 1 [true] or 0 [false]) for the match $m$, and $\Omega$ is a  
regularizer (\eg, $\Omega(w)=||w||^2$). We use the hinge loss and $L_2$ regularizer in this work. 
Finally, at test time, we associate any region $r$ with the region $s$
maximizing $z([r,s],w^*)$, where $w^*$ is the set of learned parameters.

\subsection{Similarity function and geometry kernel\label{sec:kernel}}
There are many possible
choices for the function $f$ that computes the appearance similarity
of the two regions $r$ and $s$ making up match number $m$. 
Here we assume a trainable embedding function $c$ (as will be shown later, $c$ will be the output of a CNN in our case) that outputs a $L_2$ normalized 
feature vector. 
For the appearance similarity between two regions $r$ and $s$, we then use a rectified cosine similarity:   
\begin{equation}
f(m,w)=\max(0,c(r,w)\cdot c(s,w)), \label{eq:rectdot} 
\end{equation}
that sets all negative similarity values to zero, thus making the similarity function sparser as well as insensitive to negative matches during training, with the additional benefit of giving nonnegative weights in Eq.~(\ref{eq:eq1}). 

Our geometry kernel $K_{mm'}$ records the fact that two matches (roughly)
correspond to the same offset: Concretely, we discretize the set of all possible
offsets into bins. Let us denote by $h$ the function mapping a match $m$
onto the corresponding bin $x$, we now define $g$ by
\begin{equation}
  g(m,x) =  
\begin{cases}
            1,& \text{if } h(m) = x \\
            0,         & \text{otherwise}.
\end{cases}
\end{equation}
Thus, the kernel $K_{mm'}$ simply measures whether two matches share the same
offset bin or not:   
\begin{equation}
  K_{mm'} =  
\begin{cases}
            1,& \text{if } h(m) = h(m') \\
            0,         & \text{otherwise}.
\end{cases}\label{eq:sparsekernel}
\end{equation}
In practice, $x$ runs over a grid of predefined offset values, and $h(m)$ assigns match $m$ to the nearest offset point.  
Our kernel is sparse, which greatly simplifies the computation of the score function in Eq.~(\ref{eq:s_func}): Indeed, let $B_x$ denote the set of matches associated with the bin $x$, the score function $z$ reduces to  
\begin{equation}
  z(m,w)=f(m,w)\sum_{m'\in B_{h(m)}} f(m',w).\label{eq:ss_func}\end{equation}
This trainable form of the PHM model from~\cite{CKSP15,HCSP16} can be used within Eq.~(\ref{eq:main}).

Note that since our simple geometry kernel is only dependent on matches' offsets, we obtain the same geometry term value of $\sum_{m'\in B_{h(m)}} f(m',w)$ for any match $m$ that falls into the same bin $h(m)$. This allows us to compute this geometry term value only once for each non-empty bin $x$ and then share it for multiple matches in the same bin. This sharing makes computing $z$ several times faster in practice.\footnote{If the geometry kernel is dependent on something other than offsets, e.g., matches' absolute position or their neighborhood structure, this sharing is not possible.} 

\subsection{Gradient-based learning}

\begin{figure*}
  \centering
  \includegraphics[width=.8\linewidth]{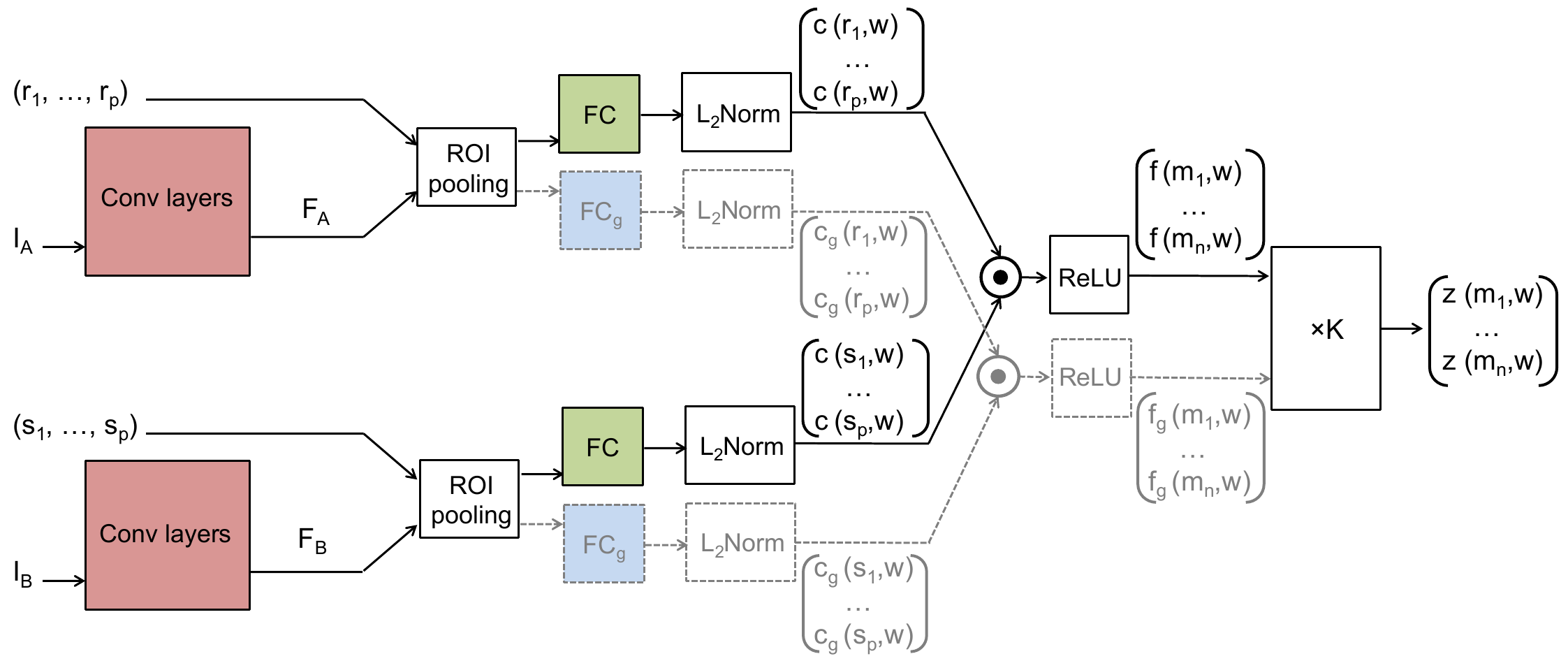}
\caption{The SCNet architectures. Three variants are proposed: SCNet-AG, SCNet-A, and SCNet-AG+. The basic architecture, SCNet-AG, is drawn in solid lines. Colored boxes represent layers with learning parameters and the boxes with the same color share the same parameters. ``$\times K$'' denotes the voting layer for geometric scoring. A simplified variant, SCNet-A, learns appearance information only by making the voting layer an identity function. An extended variant, SCNet-AG+, contains an additional stream drawn in dashed lines. SCNet-AG learns a single embedding $c$ for both appearance and geometry, whereas SCNet-AG+ learns an additional and separate embedding $c_g$ for geometry. See text for details. (Best viewed in color.)}
\label{fig:minsuarch}
\end{figure*}

The feature embedding function $c(m,w)$ in the model above can be 
implemented by any differentiable architecture, for example a
CNN-based one, and the score function $z$ can be learned using stochastic gradient descent. 
Let us now consider the problem of minimizing the objective function
$E(w)$ defined by Eq.~(\ref{eq:main}).\footnote{We take $\Omega(w)=0$ for
simplicity in this section, but tackling a nonzero regularizer is
easy.} This requires computing the gradient with respect to $w$ of the function $z$:
\begin{eqnarray}
\nabla z(m,w)=[\sum_{m'\in D} K_{mm'} f(m',w)]\nabla f(m,w) \nonumber\\
+f(m,w)\sum_{m'\in D} K_{mm'} \nabla f(m',w).\label{eq:gs_func}
\end{eqnarray}
Denoting by $n$ the size of $D$, this involves $n$ evaluations of both $f$ and $\nabla f$.  Computing the full gradient of $E$ thus requires at most $n^2$ evaluations of both $f$ and $\nabla f$, which becomes computationally intractable when $n$ is large enough.
The score function of Eq.~(\ref{eq:ss_func}) with the sparse kernel of Eq.~(\ref{eq:sparsekernel}), however, greatly reduces the gradient computation:\vspace{-0.3cm}
\begin{eqnarray}
\nabla z(m,w)=[\sum_{m'\in B_{h(m)}} f(m',w)]\nabla f(m,w) \nonumber \\
+f(m,w)\sum_{m'\in B_{h(m)}} \nabla f(m',w).
\end{eqnarray}
Note that computing the gradient for match $m$ involves only a small set of matches falling into the same offset bin $h(m)$. 

\section{SCNet architecture}

Among many possible architectures implementing the proposed model, 
we propose using a convolutional neural network (CNN), dubbed {\em SCNet}, that efficiently processes  regions and learns our matching model. Three variants, SCNet-AG, SCNet-A, SCNet-AG+, are illustrated in Fig.~\ref{fig:minsuarch}. 

In each case, SCNet takes as input two images $I_A$ and $I_B$, and maps them onto
feature maps $F_A$ and $F_B$ by CNN layers. Given region proposals
$(r_1,\ldots,r_p)$ and $(s_1,\ldots,s_p)$ for the two images, parallel ROI
pooling layers~\cite{girshickICCV15fastrcnn,kaiming14ECCV} extract feature maps of the same size for each proposal. This is an efficient architecture that shares convolutional layers over all region proposals. 

\vspace{-0.3cm}
\paragraph{SCNet-AG.} The proposal features are fed into a fully-connected layer, mapped onto feature embedding vectors, and normalized into unit feature vectors $c(r_i,w)$ and $c(s_j,w)$, associated with the regions $r_i$ and
$s_j$ of $I_A$ and $I_B$, respectively. The value of
$f(m,w)$ for the match $m$ associated with regions $r_i$ and $s_j$ is computed as the
rectified dot product of $c(r_i)$ and $c(s_j)$ (Eq.~(\ref{eq:rectdot})), which defines the appearance similarity  $f(m,w)$ for match $m$.
Geometric consistency is enforced with the kernel described in Sec.~\ref{sec:kernel}, using a voting layer, denoted as ``$\times K$'', that computes score $z(m,w)$ from the appearance similarity and geometric consensus of proposals. Finally, matching is performed by identifying the maximal $z(m,w)$ scores, using both appearance and geometric similarities.

\vspace{-0.3cm}
\paragraph{SCNet-A.} We also evaluate a similar architecture without the geometry term. This architecture drops the voting layer (denoted by $\times K$ in Fig.~\ref{fig:minsuarch}) from SCNet-AG, directly using $f(m,w)$ as a score function. This is similar to the universal correspondence network (UCN)~\cite{UCN16}. The main differences are the use of object proposals and the use of a different loss function.    

\vspace{-0.3cm}
\paragraph{SCNet-AG+.} Unlike SCNet-AG, which learns a single embedding $c$ for both appearance and geometry, SCNet-AG+ learns an additional and separate embedding $c_g$ for geometry that is implemented by an additional stream in the SCNet architecture (dashed lines in Fig.~\ref{fig:minsuarch}). This corresponds to a variant of Eq. (\ref{eq:ss_func}), as follows:   
\begin{equation}
  z^{+}(m,w)=f(m,w)\sum_{m'\in B_{h(m)}} f_g(m',w),\label{eq:ss_func_p}\end{equation}
where $f_g$ is the rectified cosine similarity computed by $c_g$. Compared to the original score function, this variant allows the geometry term to learn a separate embedding function for geometric scoring. This may be beneficial particularly when a match's contribution to geometric scoring needs to be different from the appearance score. For example,  a match of rigid object parts (wheel of cars) may contribute more to geometric scoring than that of deformable object parts (leg of horses). The separate similarity function $f_g$ allows more flexibility in learning the geometric term.  

\vspace{-0.3cm}
\paragraph{Implementation details.}

We use the VGG16~\cite{SiZi2014vgg} model that consists of a set of convolutional layers with $3 \times 3$ filters, a ReLU layer and a pooling layer. We find that taking the first 4 convolutional layers is a good trade-off for our semantic feature extraction purpose without loosing localization accuracy. These layers output features with $512$ channels. For example, if the net takes input of $224\times 224 \times 3$ images, the convolutional layers produce features with the size of $14\times 14 \times 512$. For the ROI pooling layer, we choose a $7 \times 7$ filter following the fast R-CNN architecture~\cite{girshickICCV15fastrcnn}, which produces a feature map with size of $7\times 7 \times 512$ for each proposal. 
To transform the feature map for each proposal into a feature vector, we use the $FC$ layer with a size of $7 \times 7 \times 512 \times 2048$. The $2048$ dimensional feature vector associated with each proposal are then fed into the $L_2$ normalization layer, followed by the dot product layer, ReLU, our geometric voting layer, and loss layer. 
The convolutional layers are initialized by the pretrained weights of VGG16 and the fully connected layers have random initialization. 
We train our SCNet by mini-batch SGD, with learning rate $0.001$, and weight decay $0.0005$.
During training, each mini-batch arises from a pair of images associated with a number of proposals. In our implementation, we generated $500$ proposals for each image, which leads to $500 \times 500$ potential matches.

For each mini-batch, we sample matches for training as follows. (1) Positive sampling: For a proposal $r_i$ in $I_A$, we are given its ground truth match $r'_i$ in $I_B$. We pick all the proposals $s_j$ in $I_B$ with $\text{IoU}(s_j, r'_i) > T_{pos}$ to be positive matches for $r_i$. (2) Negative sampling: Assume we obtain $k$ positive pairs w.r.t $r_i$. We also need to have $k$ negative pairs w.r.t $r_i$. To achieve this, we first find the proposals $s_t$ in $I_B$ with $IoU(s_t, r'_i) < T_{neg}$. Assuming $p$ proposals satisfying the IoU constraint, we find the proposals with top $k$ appearance similarity with $r_i$ among those $p$ proposals.
In our experiment, we set $T_{pos} = 0.6$, and $T_{neg} = 0.4$.

\section{Experimental evaluation}
In this section we present experimental results and analysis. Our code and models will be made available online: {\small \url{http://www.di.ens.fr/willow/research/scnet/}}. 
\subsection{Experimental details}

\paragraph{Dataset.}
We use the PF-PASCAL dataset that consists of 1300 image pairs selected from PASCAL-Berkeley keypoint annotations\footnote{http://www.di.ens.fr/willow/research/proposalflow/} of 20 object classes. Each pair of images in PF-PASCAL share the same set of non-occluded keypoints. We divide the dataset into 700 training pairs, 300 validation pairs, and 300 testing pairs. The image pairs for training/validation/testing are distributed proportionally to the number of image pairs of each object class. 
In training, we augment the data into a total of 1400 pairs by horizontal mirroring. We also test our trained models with the PF-WILLOW dataset~\cite{HCSP16}, Caltech-101~\cite{fei2006one} and PASCAL Parts~\cite{zhou2015flowweb} to further validate a generalization of the models.

\vspace{-0.3cm}
\paragraph{Region proposal.}
Unless stated otherwise, we choose to use the method of Manen~\etal (RP)~\cite{manen2013prime}. The use of RP proposals is motivated by the superior result reported in \cite{HCSP16}, which is verified once more by our evaluation. In testing we use 1000 proposals for each image as in~\cite{HCSP16}, while in training we use 500 proposals for efficiency.


\vspace{-0.3cm}
\paragraph{Evaluation metric.}
We use three metrics to compare the results of SCNet to other methods. First, we use the probability of correct keypoint (PCK) \cite{yang2013articulated}, which measures the precision of dense flow at sparse keypoints of semantic relevance. It is calculated on the Euclidean distance $d(\phi(p), p^*)$ between a warped keypoint $\phi(p)$ and ground-truth one $p^*$\footnote{To better take into account the different sizes of images, we normalize the distance by dividing by the diagonal of the warped image, as in \cite{UCN16}}. Second, we use the probability of correct regions (PCR) introduced in \cite{HCSP16} as an equivalent of the the PCK for region based correspondence. PCR measures the precision of a region matching between region $r$ and its correspondent $r^*$ on the intersection over union (IoU) score $1-\text{IoU}(\phi(r),r^*)$. Both metrics are computed against a threshold $\tau$ in $[0,1]$ and we measure {PCK}@$\tau$ and {PCR}@$\tau$ as the percentage correct below $\tau$. Third, we capture the quality of matching proposals by the mean IoU of the top $k$ matches (mIoU@$k$). Note that these metrics are used to evaluate two different types of correspondence. Indeed, PCK is an evaluation metric for dense flow field, whereas PCR and mIoU@$k$ are used to evaluate region-based correspondences~\cite{HCSP16}.

\begin{figure*}[t]
  \centering
     \tabcolsep=0.02cm
   \renewcommand{\arraystretch}{0.25}
   \begin{tabular}{
         >{\centering\arraybackslash} m{0.25\textwidth}
         >{\centering\arraybackslash} m{0.25\textwidth}
         >{\centering\arraybackslash} m{0.25\textwidth}
         >{\centering\arraybackslash} m{0.25\textwidth}}
      \includegraphics[width=1\linewidth]{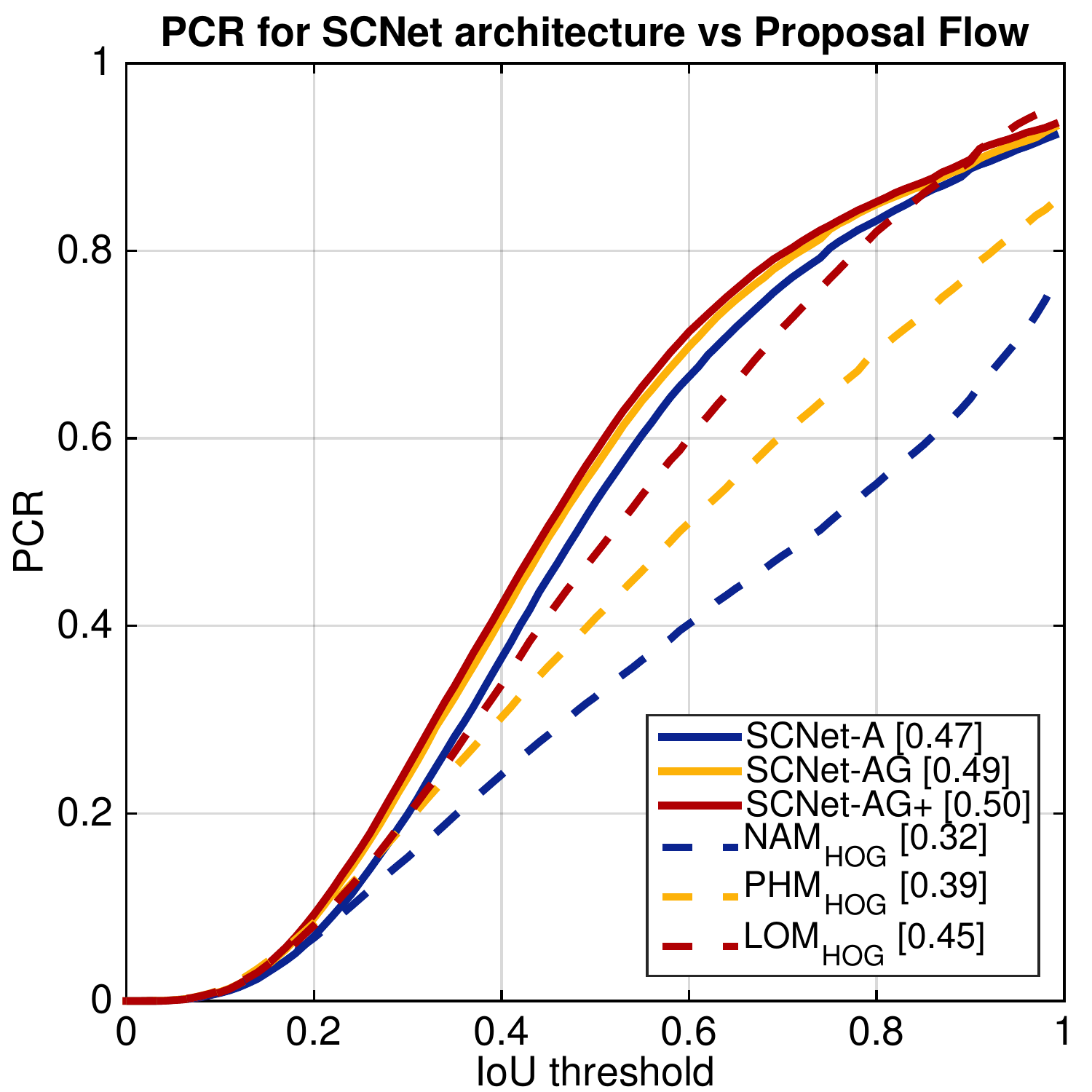} &
      \includegraphics[width=1\linewidth]{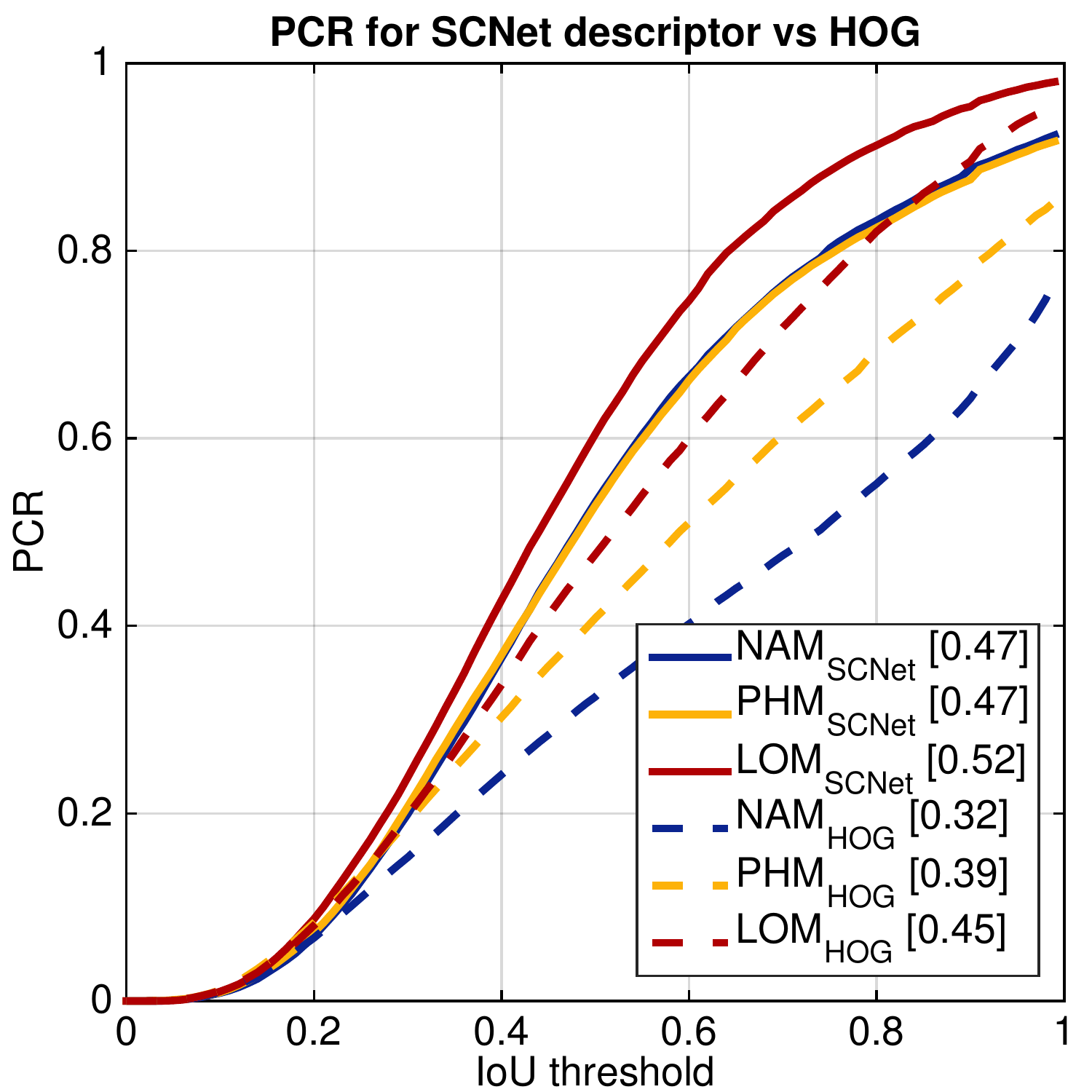} &
      \includegraphics[width=1\linewidth]{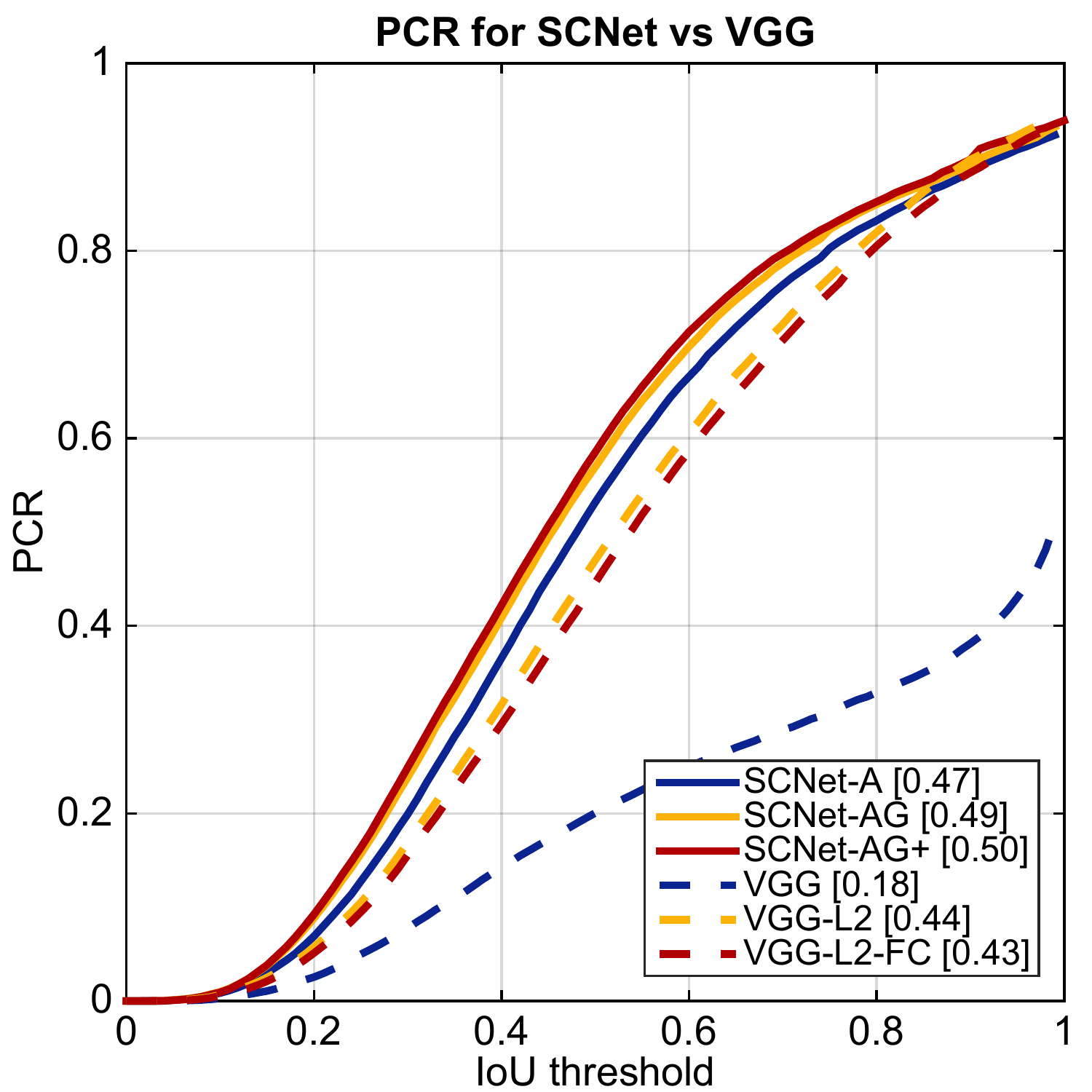} &
      \includegraphics[width=1\linewidth]{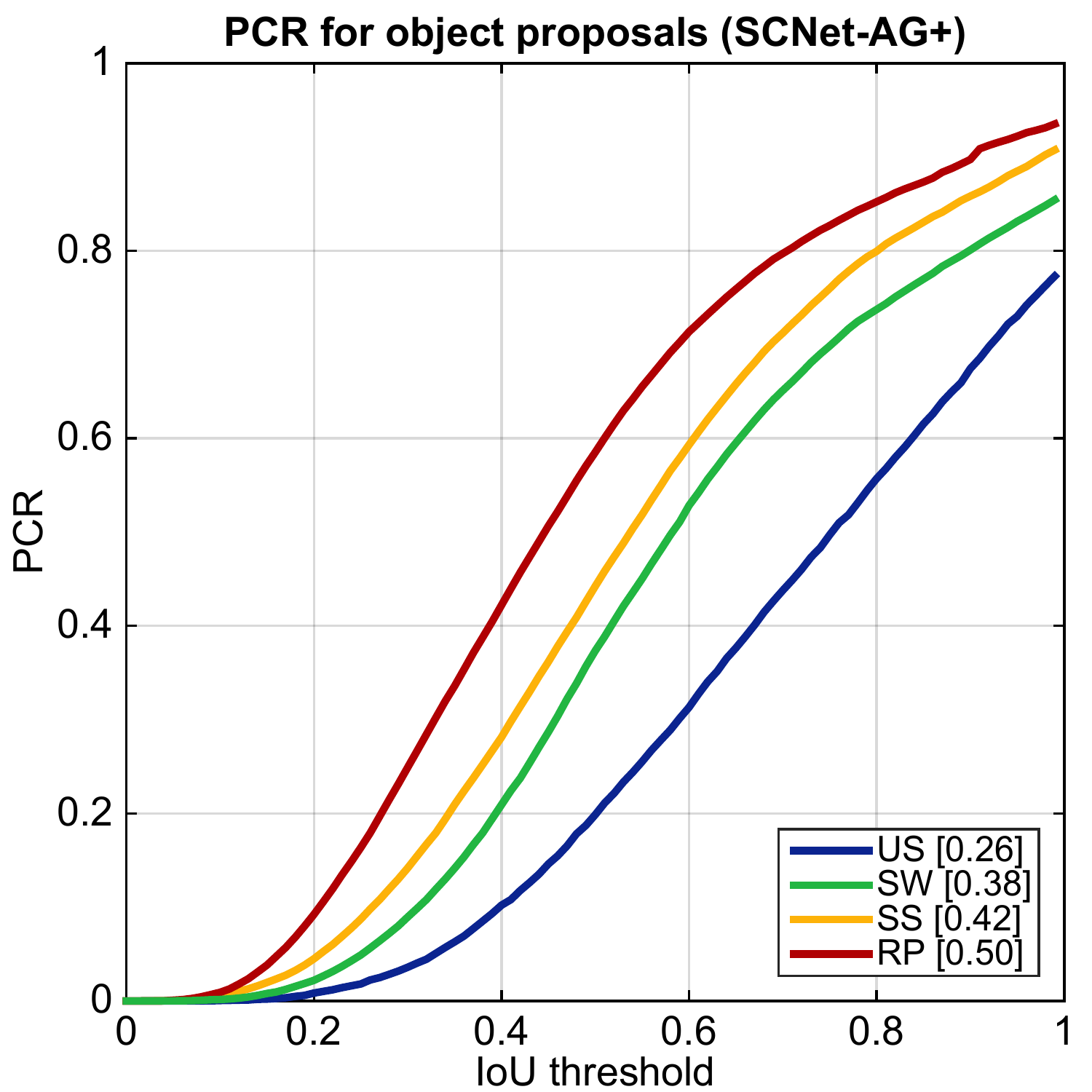}\\
      \includegraphics[width=1\linewidth]{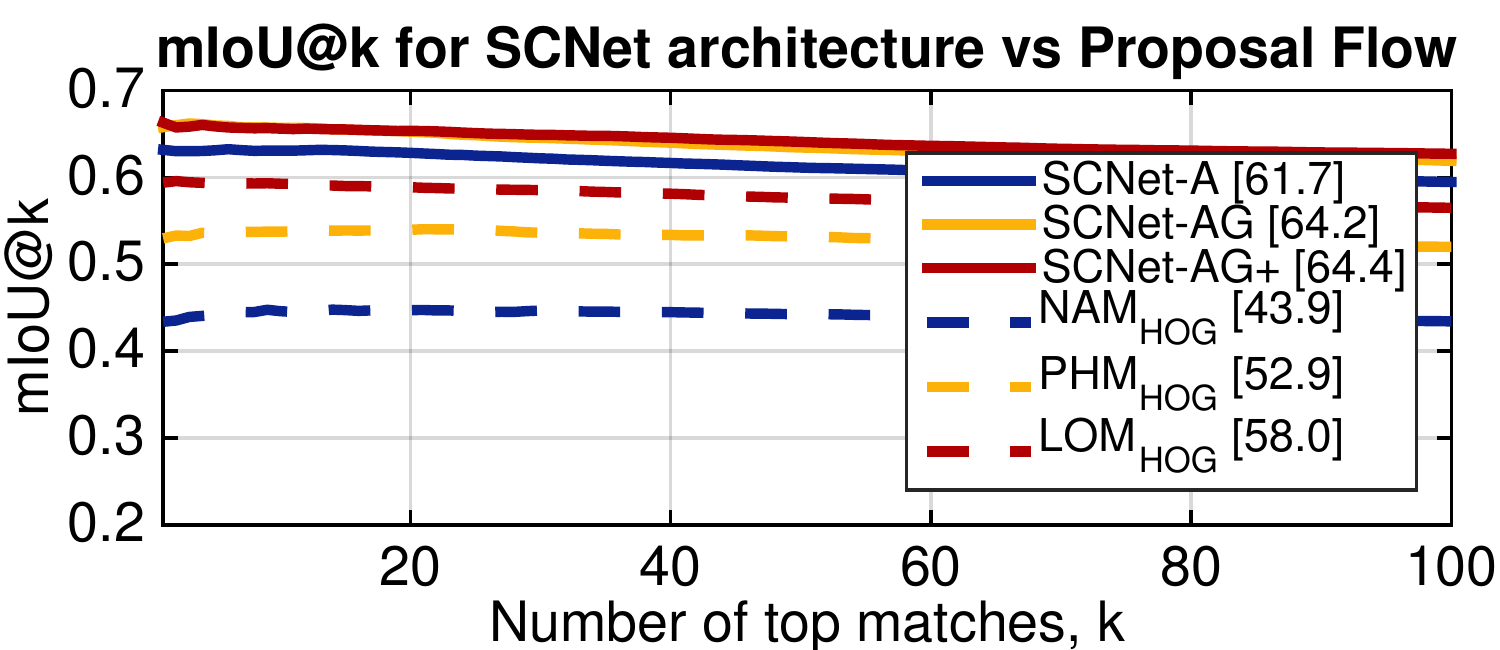} &
      \includegraphics[width=1\linewidth]{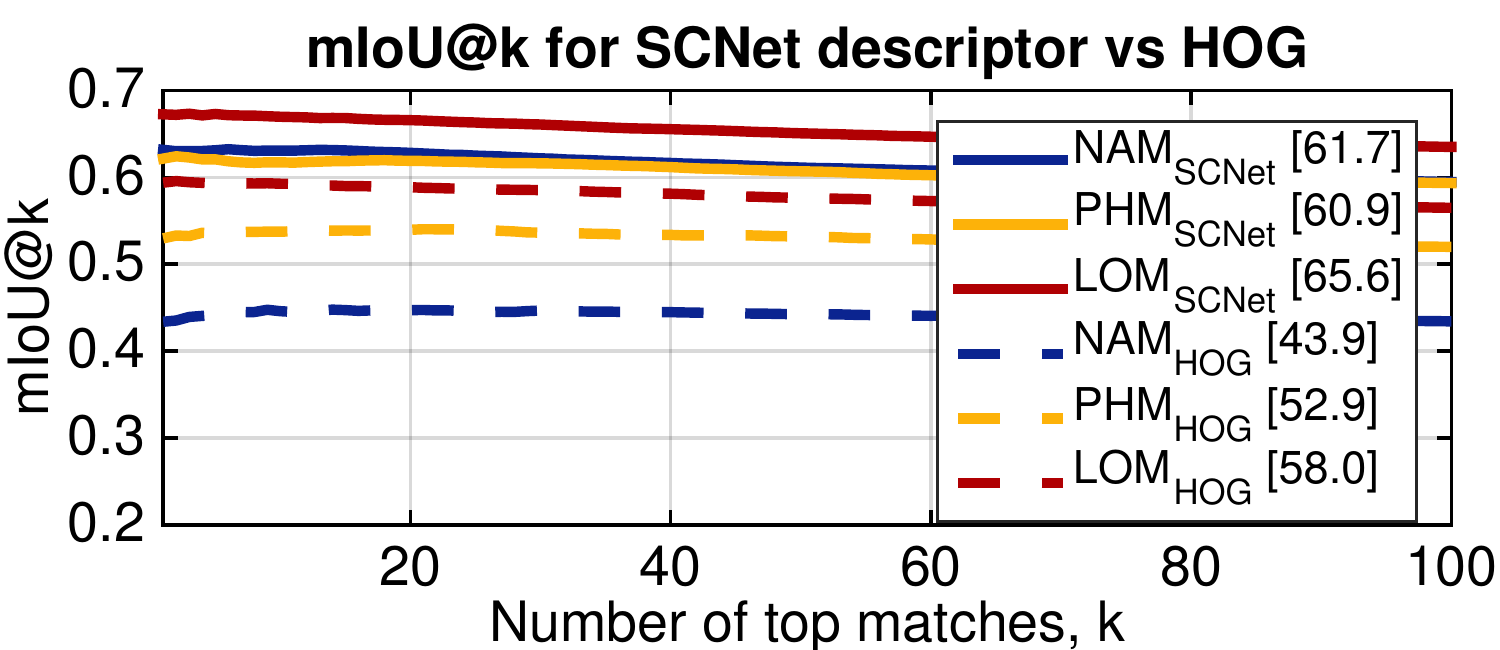} &
      \includegraphics[width=1\linewidth]{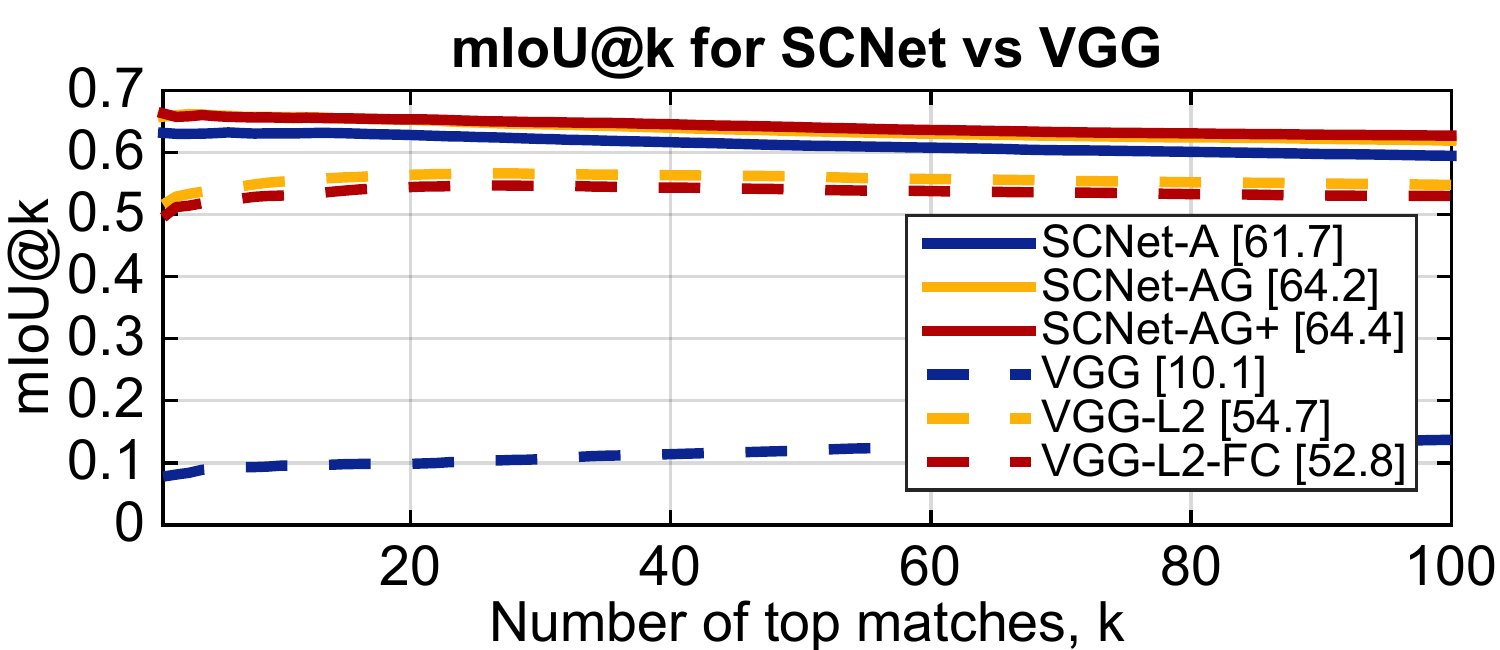} &
      \includegraphics[width=1\linewidth]{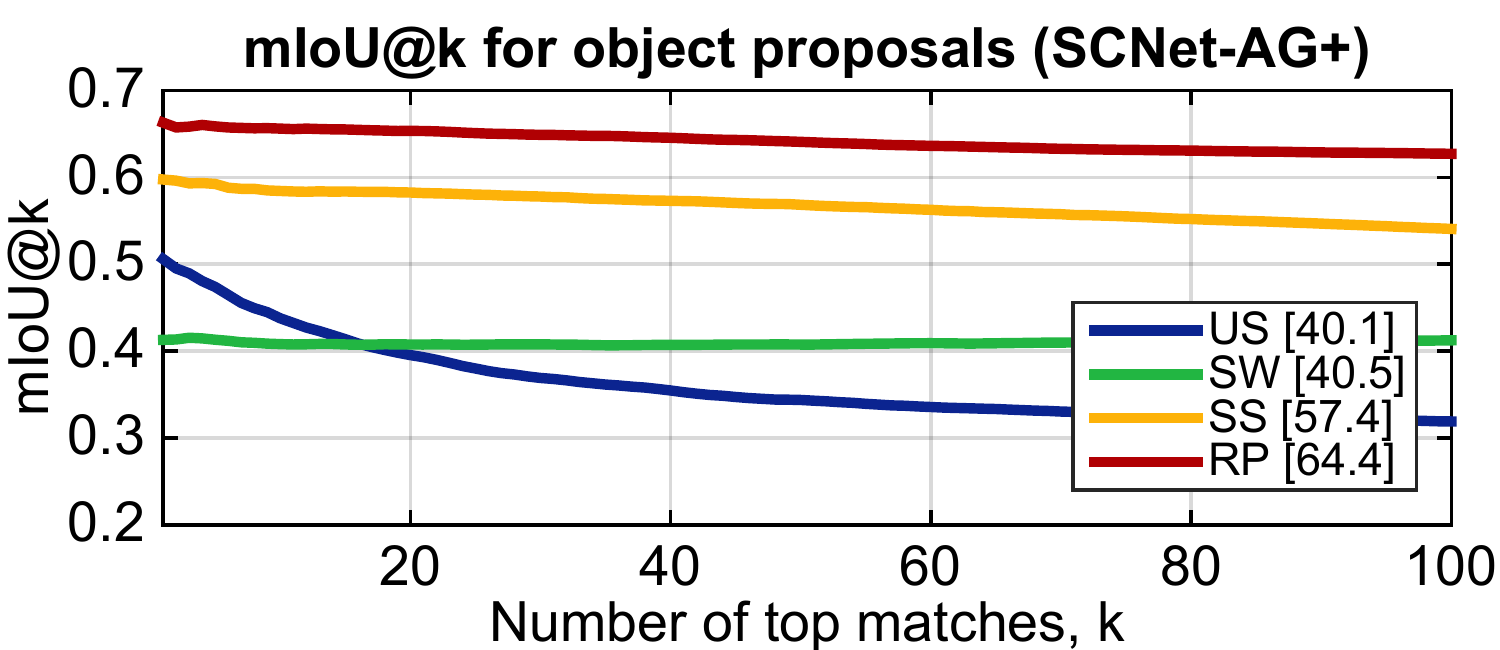}\\
      (a) & (b) & (c) & (d)\\
   \end{tabular}
   \caption{(a) Performance of SCNet on PF-PASCAL, compared to Proposal Flow methods~\cite{HCSP16}. (b) Performance of SCNet and HOG descriptors on PF-PASCAL, evaluated using Proposal Flow methods~\cite{HCSP16}. (c) Comparison to ImageNet-trained baselines. (d) Comparison of different proposals. PCR and mIoU@$k$ plots are shown at the top and bottom, respectively. AuC is shown in the legend. (Best viewed in pdf.)}
   \label{fig:PCR_mIoU_PF_PASCAL}
\end{figure*}

\subsection{Proposal flow components}


We use the PF-PASCAL dataset to evaluate region matching performance. This setting allows our method to be tested against three other methods in \cite{HCSP16}: NAM, PHM and LOM. NAM finds correspondences using handcrafted features only. PHM and LOM additionally consider global and local geometric consistency, respectively, between region matchings. We also compare our SCNet-learned feature against whitened HOG~\cite{dalal2005histograms}, the best performing handcraft feature of \cite{HCSP16}. Experiments on the PF-WILLOW dataset~\cite{HCSP16} showed similar results with the ones on the PF-PASCAL dataset. For details, refer to our project webpage.

\begin{figure*}[t]
  \centering
     \tabcolsep=0.05cm
   \renewcommand{\arraystretch}{0.5}
   \begin{tabular}{
         >{\centering\arraybackslash} m{0.23\textwidth}
         >{\centering\arraybackslash} m{0.23\textwidth}
         >{\centering\arraybackslash} m{0.23\textwidth}
         >{\centering\arraybackslash} m{0.23\textwidth}}
      \includegraphics[width=1\linewidth]{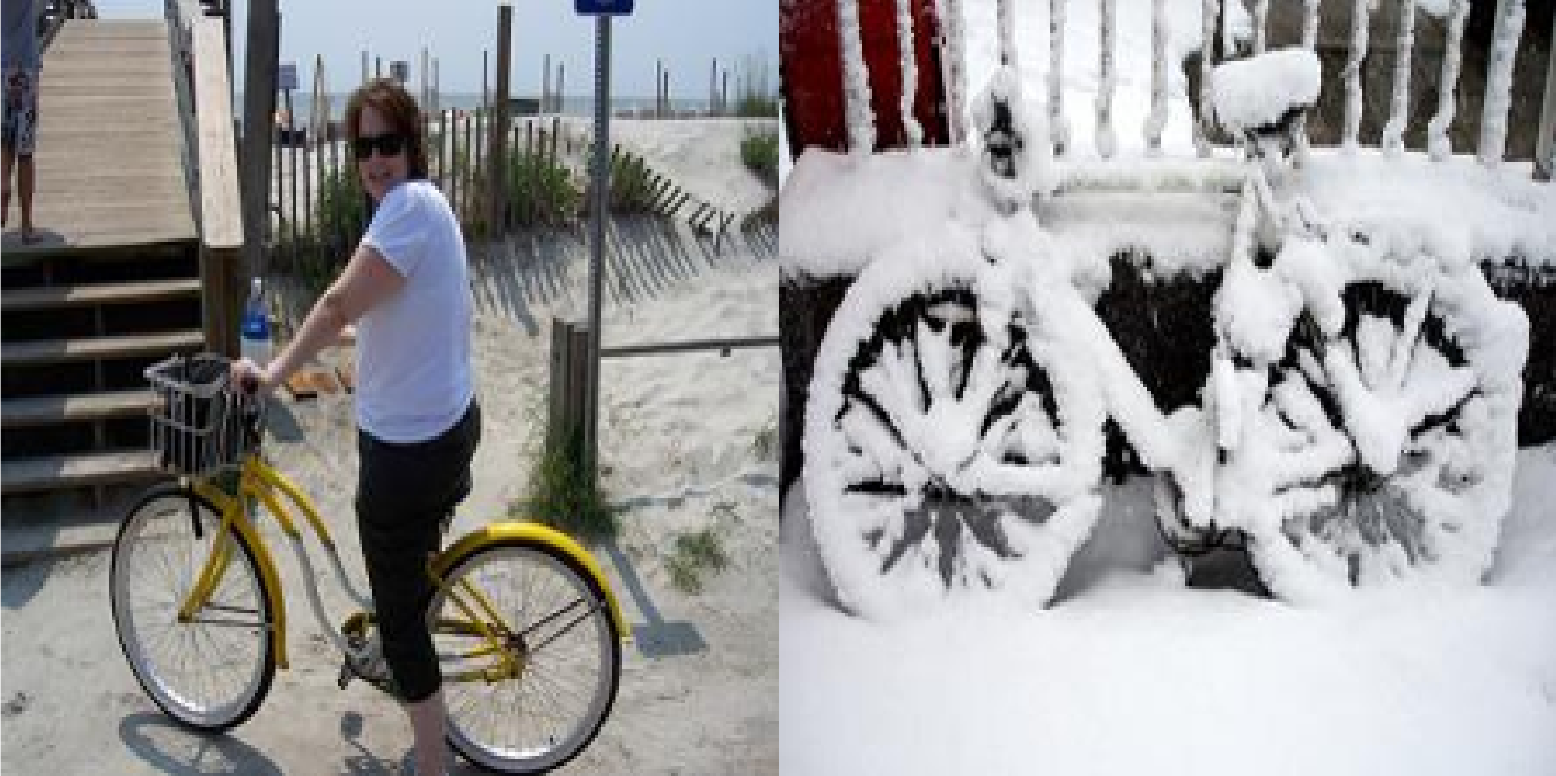} &
      \includegraphics[width=1\linewidth]{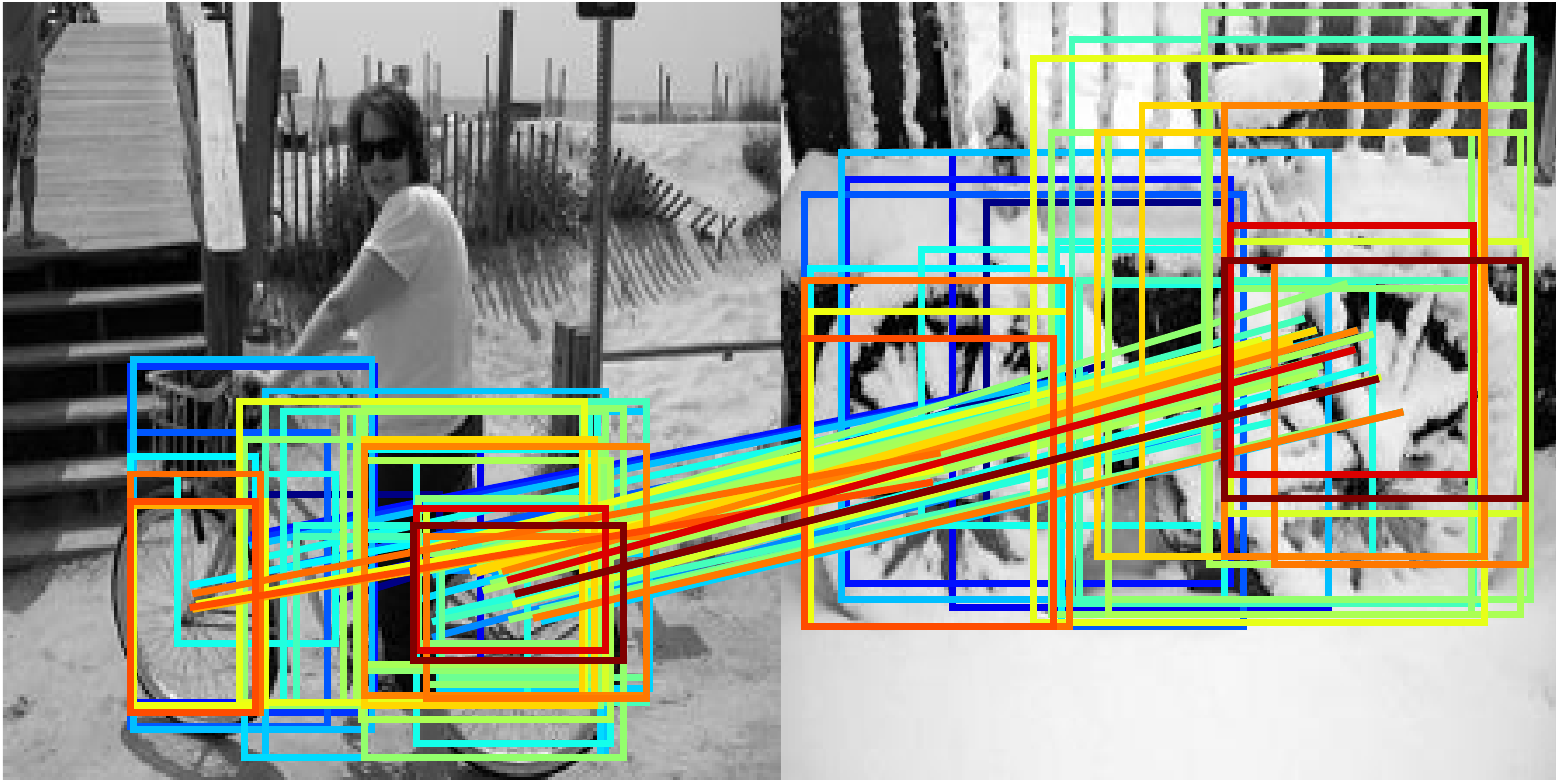} &
      \includegraphics[width=1\linewidth]{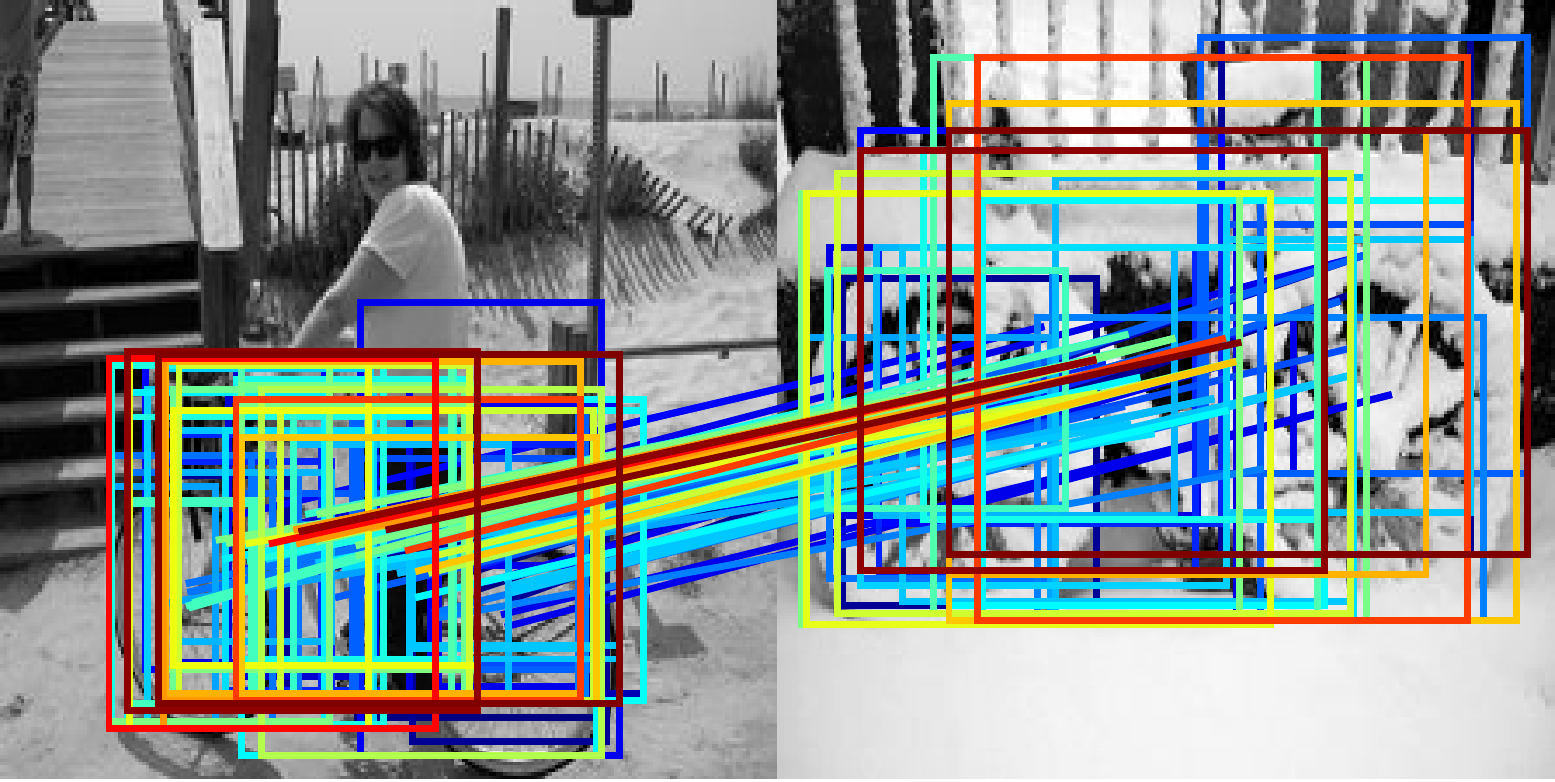} &
      \includegraphics[width=1\linewidth]{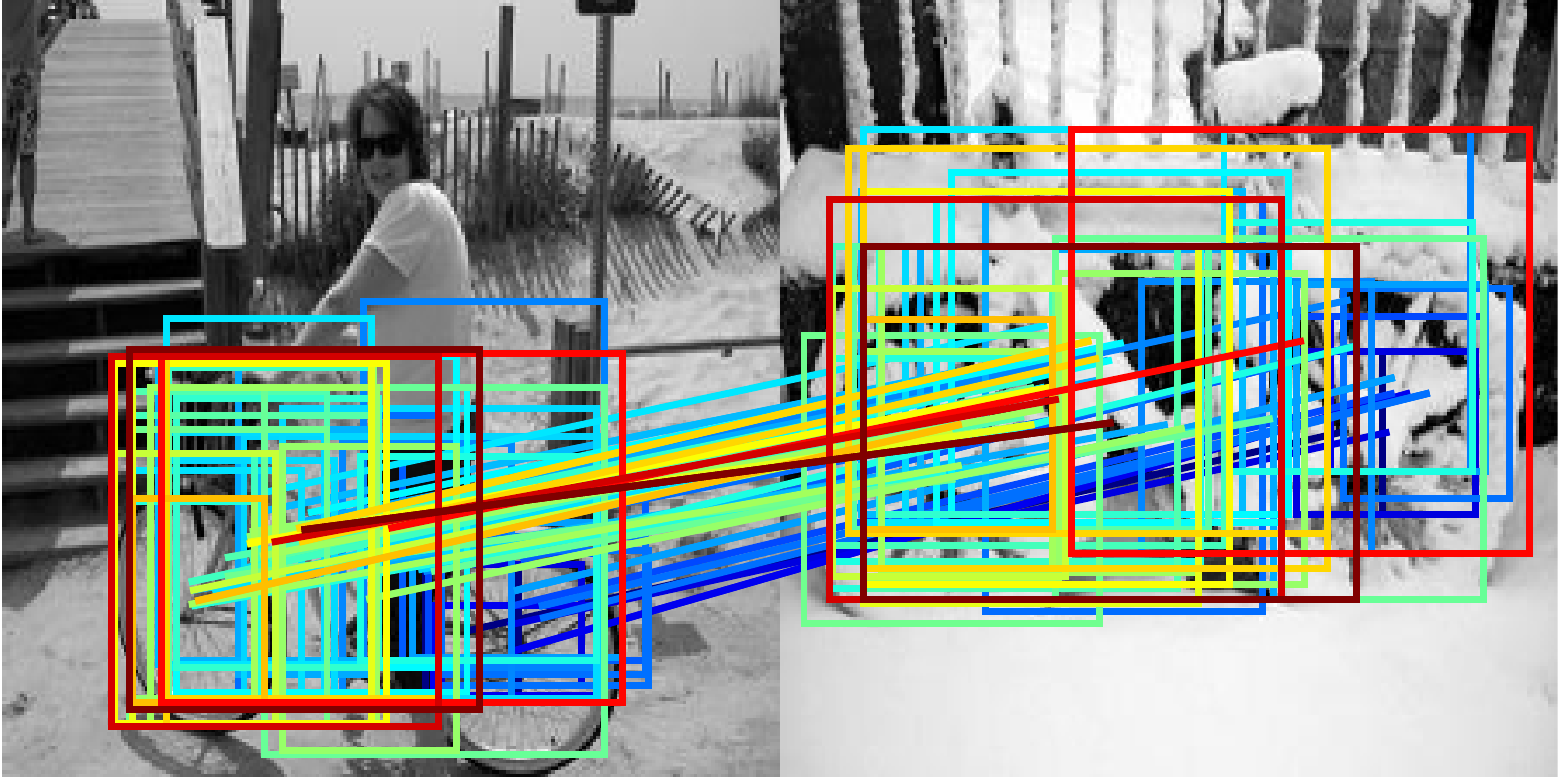} \\
       $bike$ image pair&
       NAM$_{\rm HOG}$   [37]&
       SCNet-A  [104]&
       SCNet-AG+ [107]\\
      \includegraphics[width=1\linewidth]{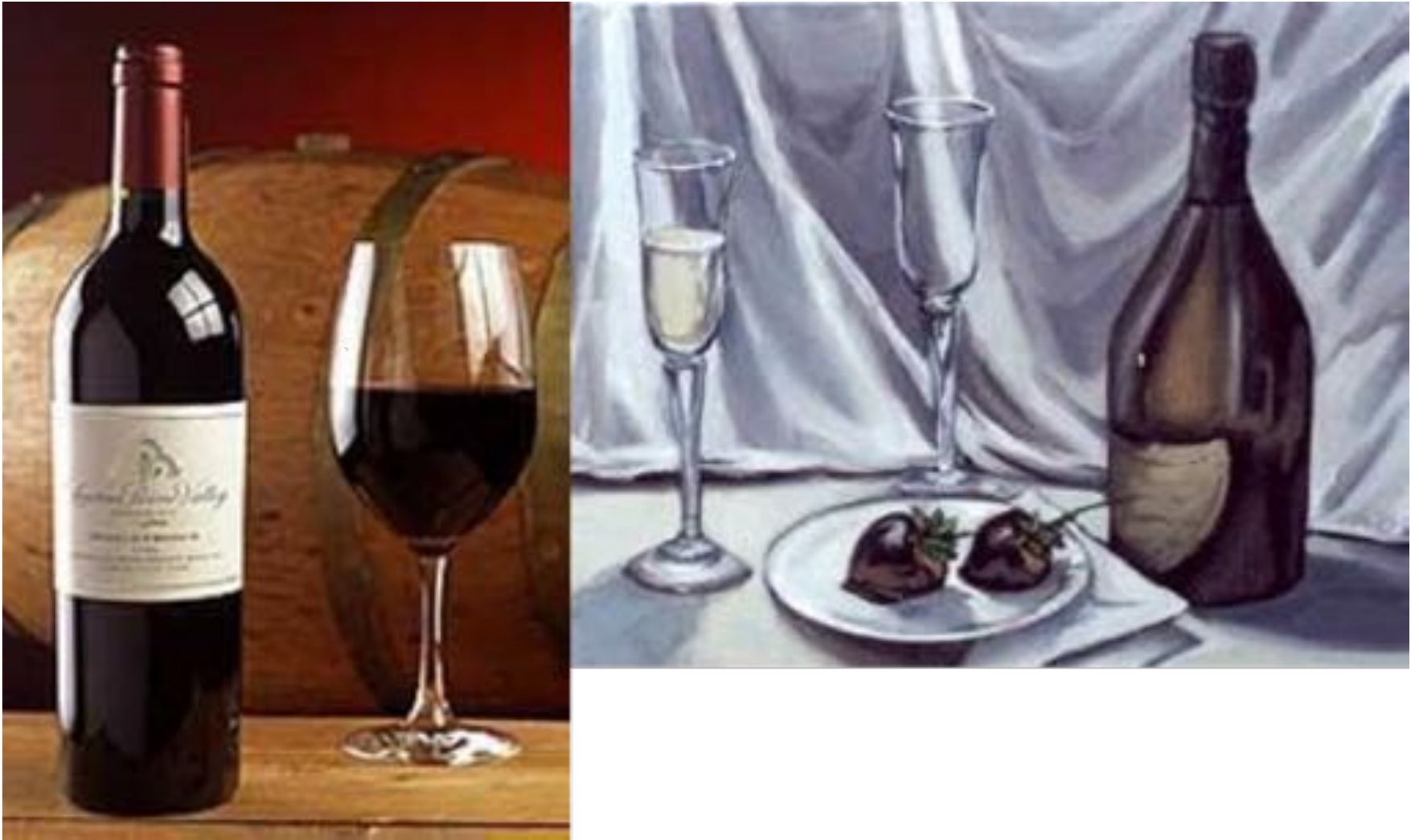} &
      \includegraphics[width=1\linewidth]{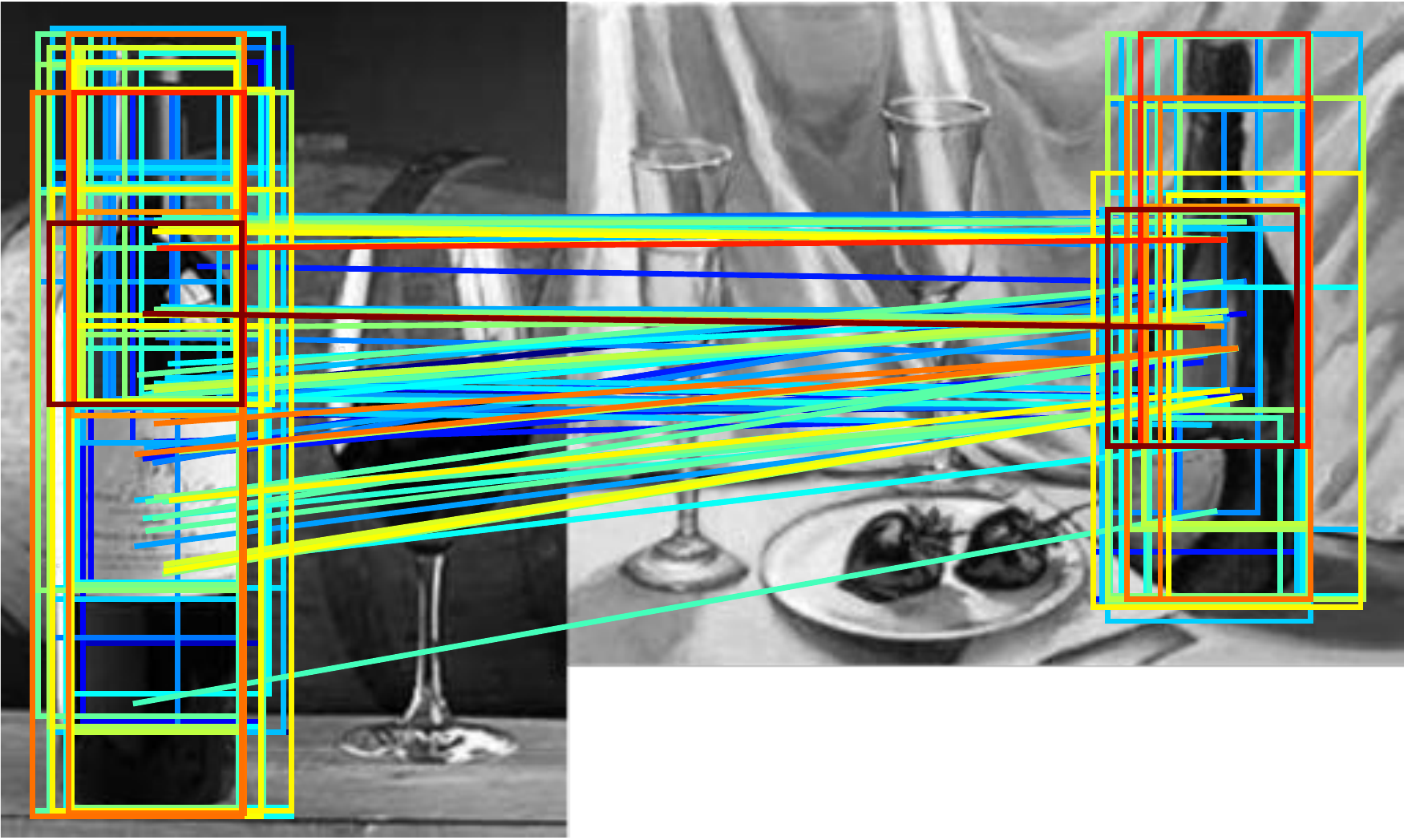} &
      \includegraphics[width=1\linewidth]{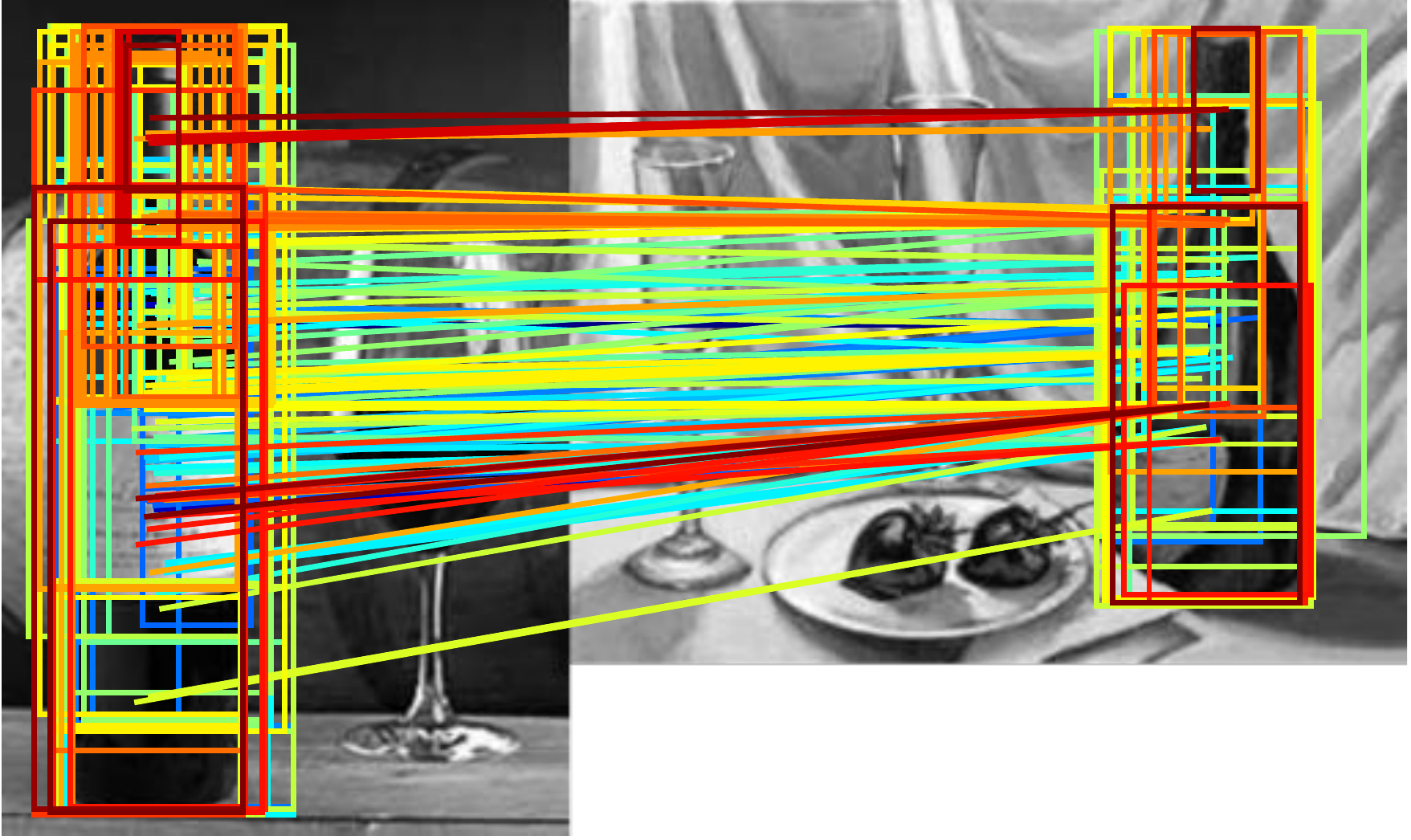} &
      \includegraphics[width=1\linewidth]{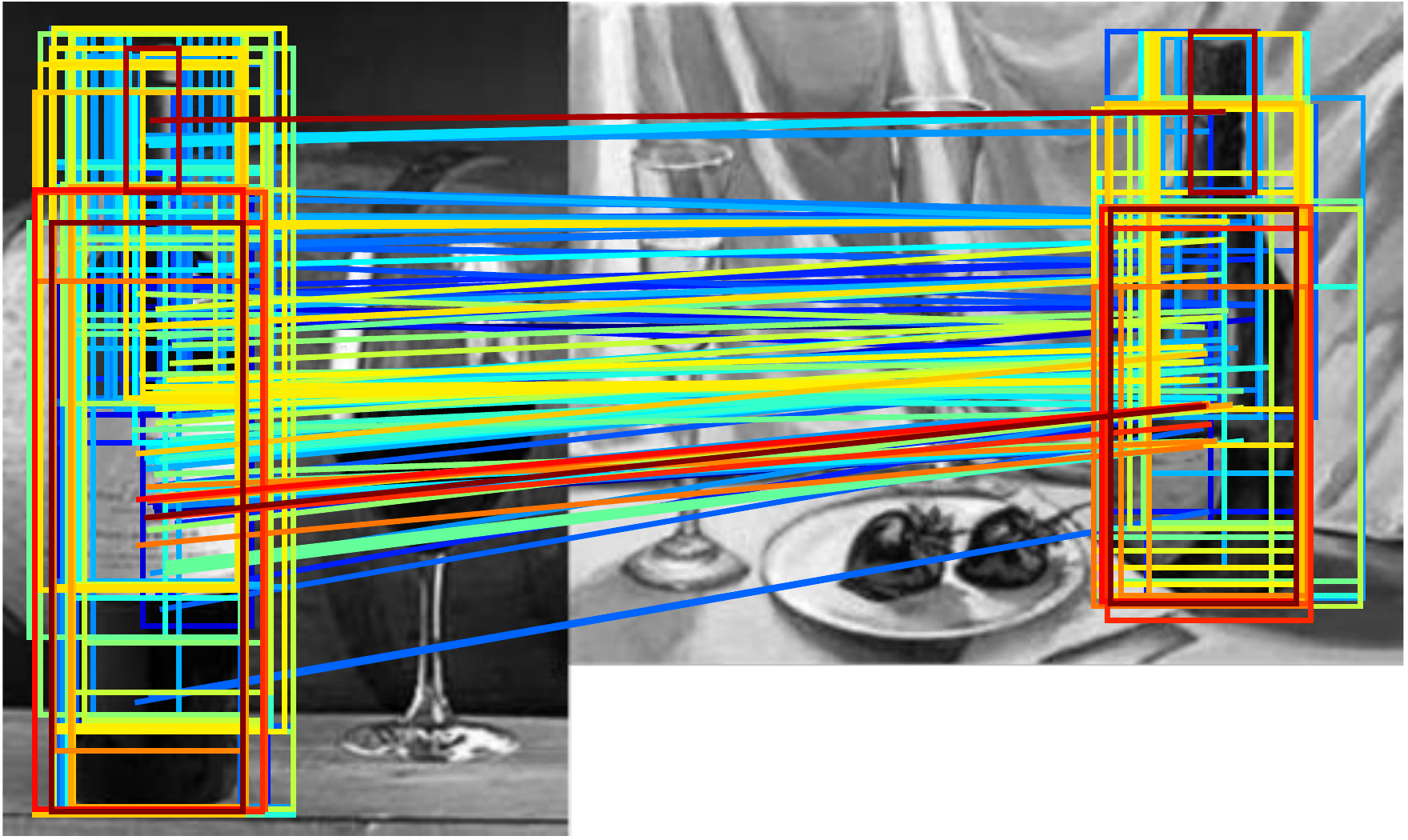} \\
       $wine\;bottle$ image pair&
       NAM$_{\rm HOG}$   [88]&
       SCNet-A  [177]&
       SCNet-AG+ [180]\\
   \end{tabular}
   \caption{Region matching examples. Numbers beside methods stand for numbers of correct matches.}
   \label{fig:qanli_PF_PASCAL}
\end{figure*}

\vspace{-0.3cm}
\paragraph{Quantitative comparison.}

Figure~\ref{fig:PCR_mIoU_PF_PASCAL}(a) compares SCNet methods with the proposal flow methods~\cite{HCSP16} on the PF-PASCAL dataset. Our SCNet models outperform the other methods that use the HOG feature. Our geometric models (SCNet-AG, SCNet-AG+) substantially outperform the appearance-only model (SCNet-A), and SCNet-AG+ slightly outperform SCNet-AG. This can also be seen from the area under curve (AuC) presented in the legend. This clearly show the effectiveness of deep learned features as well as geometric matching. In this comparison, we fix the VGG16 layer and only learn the FC layers. In our experiment, we also learned all layers including VGG 16 and the FC layers in our model (fully finetuned), but the improvement over the partially learned model was marginal. 
Figure~\ref{fig:PCR_mIoU_PF_PASCAL}(b) shows the performance of NAM, PHM, LOM matching when replacing HOG feature with our learned feature in SCNet-A.~We see that SCNet features greatly improves all the matching methods. Interestingly, LOM using SCNet feature outperforms our best performing SCNet model, SCNet-AG+. 
However, the LOM method is more than 10 times slower than SCNet-AG+: on average the method takes $0.21 s$ for SCNet-A feature extraction and $3.15 s$ for the actual matching process, whereas our SCNet-AG+ only takes $0.33 s$ in total. Most of the time in LOM is spent in computing its geometric consistency term.
We further evaluated three additional baselines using ImageNet-trained VGG (see Figure~\ref{fig:PCR_mIoU_PF_PASCAL}(c)). Namely (i) VGG: We directly use the features from ImageNet-trained VGG, followed by ROI-pooling to make the features for each proposal of the same size ($7 \times7\times 512$). We then flatten the features into vectors of dimension 175616. (ii) VGG-L2: We l2-normalize the flattened feature of (i). (iii) VGG-L2-FC: We perform a random projection from (ii) to a feature of dimension $2048$ (the same dimension with SCNet, 12.25 times smaller than (i) and (ii)) by adding a randomly initialized FC layer on top of (ii). Note that this is exactly equivalent to SCNet-A without training on the target dataset. 

\begin{table*}[ht!]
\begin{center}
\tabcolsep=0.02cm
\renewcommand{\arraystretch}{1.0}
\setlength{\tabcolsep}{.15em}
\small
\caption{Per-class PCK on PF-PASCAL at $\tau=0.1$. For all methods using object proposals, we use 1000 RP proposals~\cite{manen2013prime}.}
\label{pck:table}
\begin{tabular}{c|ccccccccccccccccccccc}
\hline
Method & aero & bike & bird & boat & bottle & bus & car & cat & chair & cow & d.table & dog & horse & moto & person & plant & sheep & sofa & train & tv & mean\\
\hline
NAM$_{\rm HOG}$~\cite{HCSP16} & 72.9 & 73.6 & 31.5 & 52.2 & 37.9 & 71.7 & 71.6 & 34.7 & 26.7 & 48.7 & 28.3 & 34.0 & 50.5 & 61.9 & 26.7&  51.7 & 66.9&  48.2 & 47.8 & 59.0 & 52.5 \\
PHM$_{\rm HOG}$~\cite{HCSP16} & 78.3 &  76.8 & 48.5 & 46.7 & 45.9 & 72.5 & 72.1 & 47.9 & 49.0 & 84.0 &  37.2 & 46.5 & 51.3 & 72.7 & 38.4 & 53.6 & 67.2 & 50.9 & 60.0 & 63.4 & 60.3 \\
LOM$_{\rm HOG}$~\cite{HCSP16} & 73.3 & 74.4 & 54.4 & 50.9 & 49.6 & 73.8 & 72.9 & 63.6 & 46.1 & 79.8 & 42.5 & 48.0  & 68.3 & 66.3 & 42.1 & 62.1 & 65.2 & 57.1 & 64.4 & 58.0 & 62.5\\
UCN~\cite{UCN16} & 64.8 & 58.7 & 42.8 & 59.6 & 47.0 & 42.2 & 61.0 & 45.6 & 49.9 & 52.0 & 48.5 & 49.5 & 53.2 & 72.7 & 53.0 & 41.4 & \bf{83.3} & 49.0 & \bf{73.0} & 66.0  & 55.6\\
SCNet-A  & 67.6 & 72.9  & 69.3 & 59.7 & \bf{74.5} &  72.7 & 73.2 & 59.5 & 51.4 & 78.2 & 39.4 & 50.1 & 67.0 & 62.1 & \bf{69.3}  & \bf{68.5} & 78.2 & 63.3 & 57.7 & 59.8  & 66.3\\
SCNet-AG & 83.9 & 81.4 & \bf{70.6} & 62.5 & 60.6 & 81.3 & 81.2 & 59.5 & 53.1 & 81.2 & \bf{62.0} & 58.7 & 65.5 & 73.3 & 51.2 & 58.3 & 60.0 & 69.3& 61.5 & \bf{80.0} & 69.7\\
SCNet-AG+ & \bf{85.5} & \bf{84.4} & 66.3 & \bf{70.8} & 57.4 & \bf{82.7} & \bf{82.3} & \bf{71.6} & \bf{54.3} & \bf{95.8} & 55.2 & \bf{59.5} & \bf{68.6} & \bf{75.0} & 56.3 & 60.4 & 60.0 & \bf{73.7} & 66.5 & 76.7 & \bf{72.2}\\
\hline 
\end{tabular}
\end{center}
\end{table*}

\vspace{-0.3cm}
\paragraph{Results with different object proposals.}
SCNet can be combined with any region proposal methods. In this experiment, we train and evaluate SCNet-AG+ on PF-PASCAL with four region proposal methods: randomized prim (RP) \cite{manen2013prime}, selective search (SS) \cite{Uijlings13ijcv}, random uniform sampling (US), and sliding window (SW). US and SW are extracted using the work of~\cite{hosang2015what}, and SW is similar to regular grid sampling used in other popular methods~\cite{kim2013deformable,liu2011sift,weinzaepfel2015deepmatching}. 
Figure~\ref{fig:PCR_mIoU_PF_PASCAL}(d)
 compares matching performance in PCR and mIoU@$k$ when using the different proposals. RP performs best, and US performs worst with a large margin. This shows that the region proposal process is an important factor for matching performance.

\begin{figure*}[htbp!]
  \centering
     \tabcolsep=0.02cm
   \renewcommand{\arraystretch}{0.25}
   \begin{tabular}{
         >{\centering\arraybackslash} m{0.15\textwidth}
         >{\centering\arraybackslash} m{0.15\textwidth}
         >{\centering\arraybackslash} m{0.15\textwidth}
         >{\centering\arraybackslash} m{0.15\textwidth}
         >{\centering\arraybackslash} m{0.15\textwidth}
         >{\centering\arraybackslash} m{0.15\textwidth}}
      \includegraphics[width=1\linewidth]{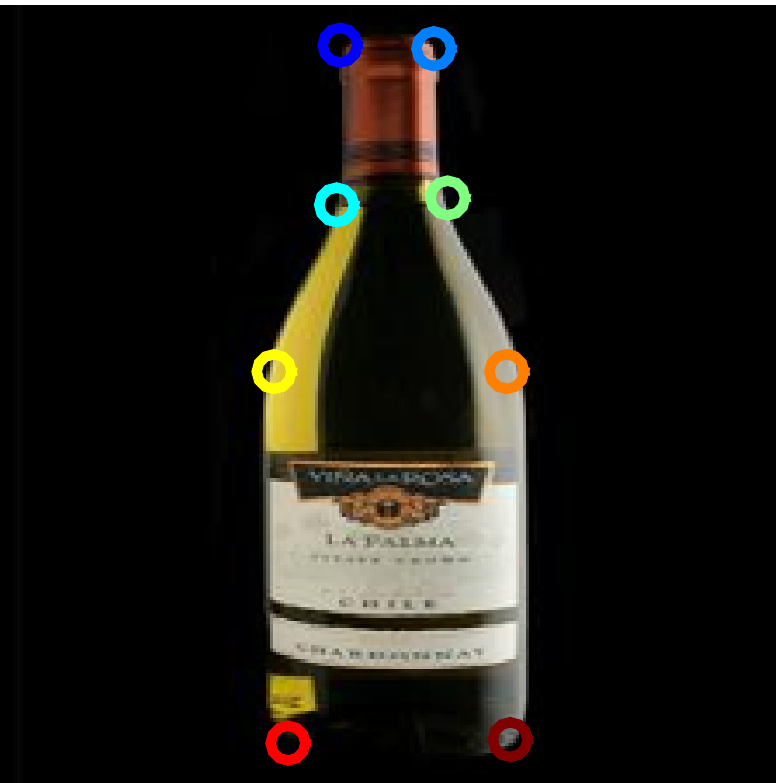} &
      \includegraphics[width=1\linewidth]{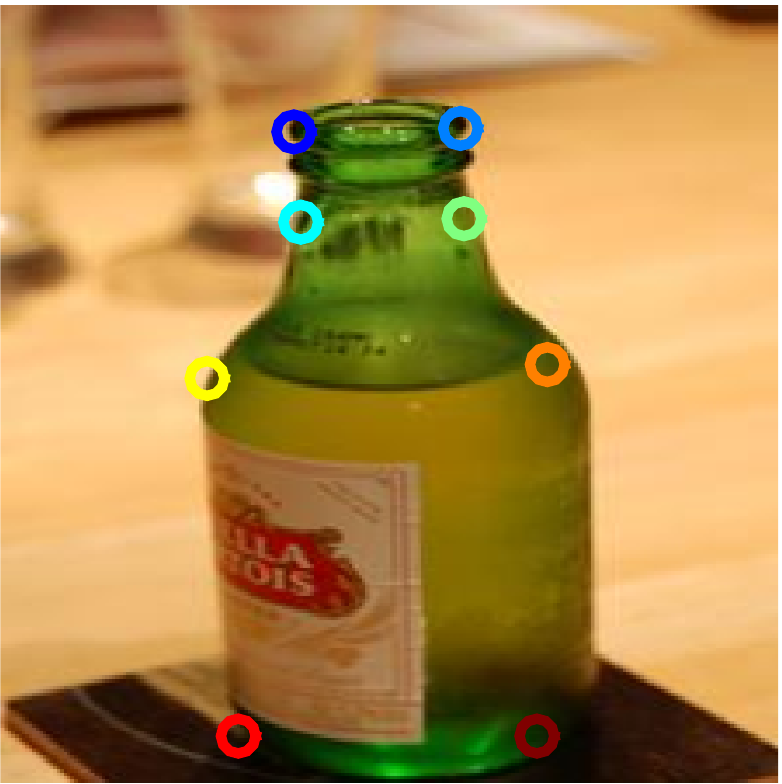} &
      \includegraphics[width=1\linewidth]{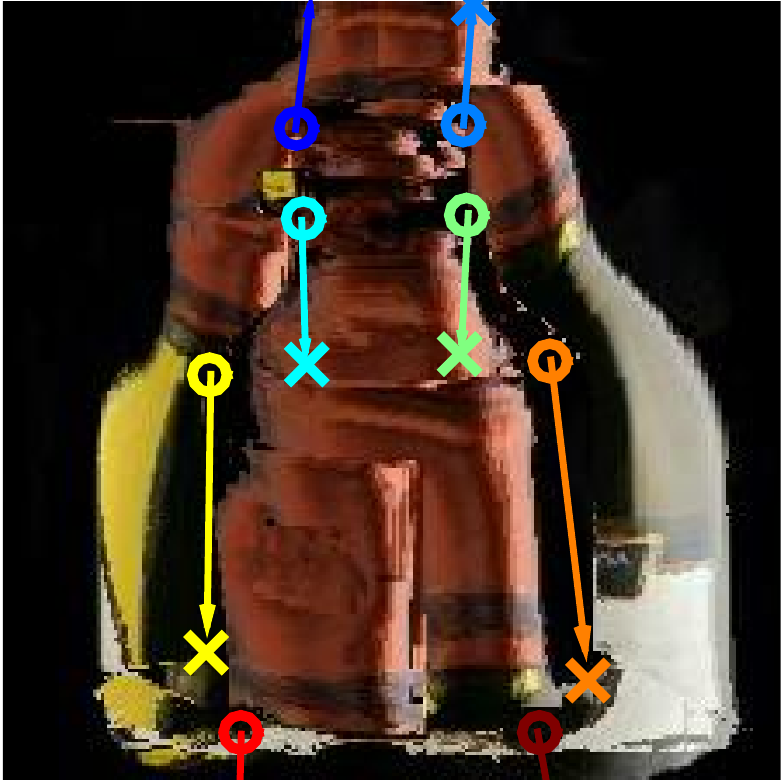} &
      \includegraphics[width=1\linewidth]{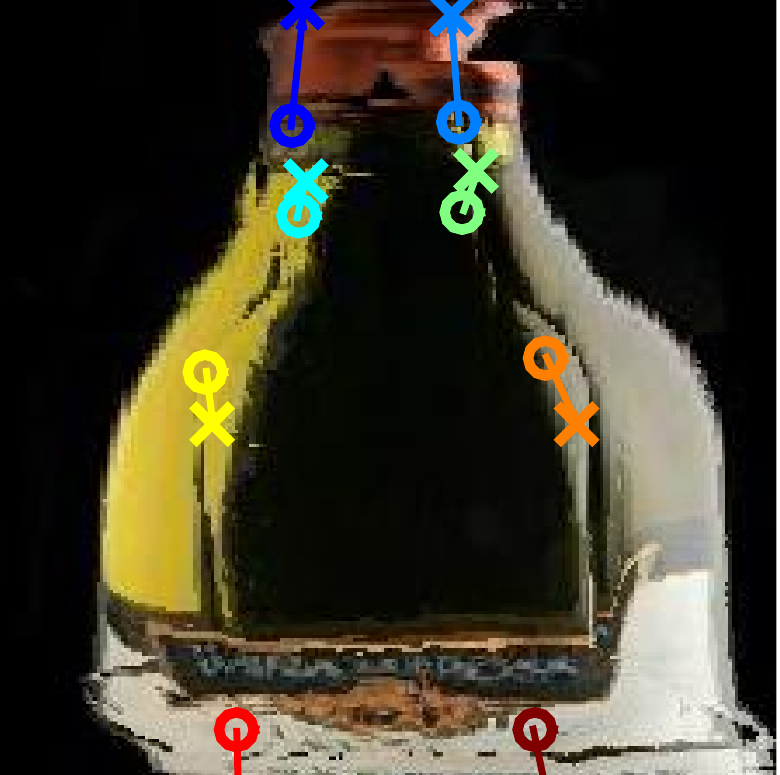} &
      \includegraphics[width=1\linewidth]{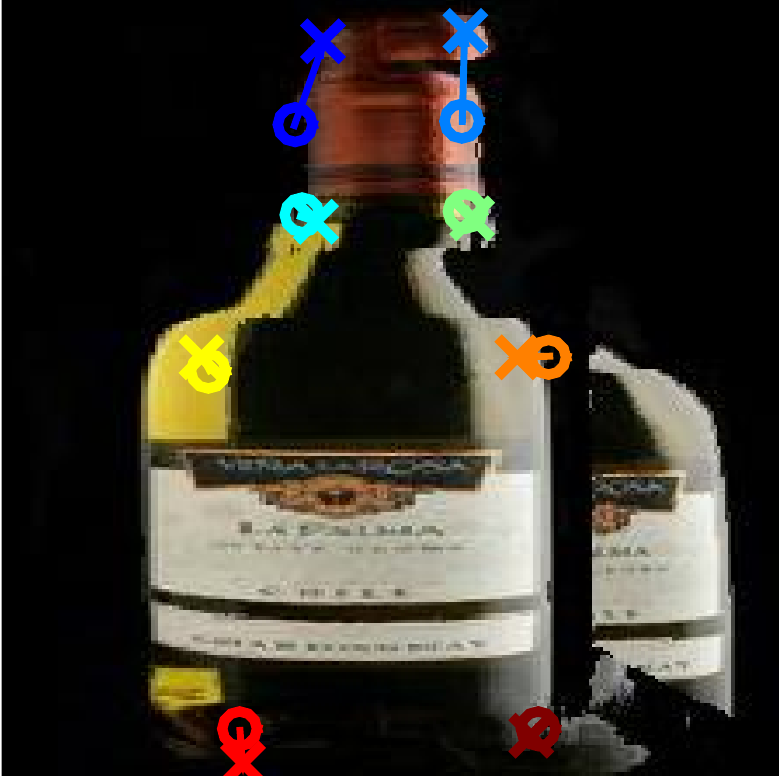} &
      \includegraphics[width=1\linewidth]{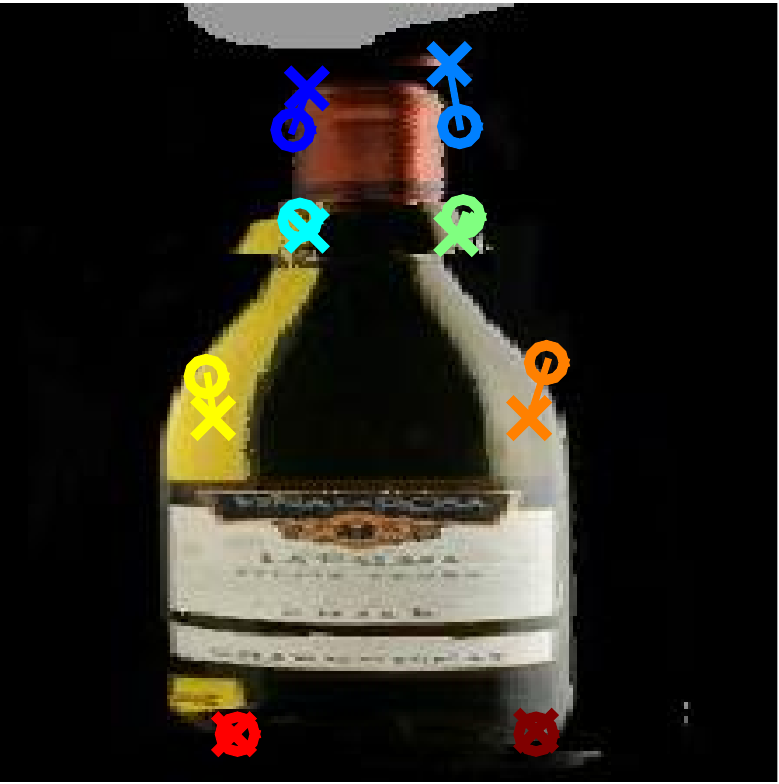}\\

      \includegraphics[width=1\linewidth]{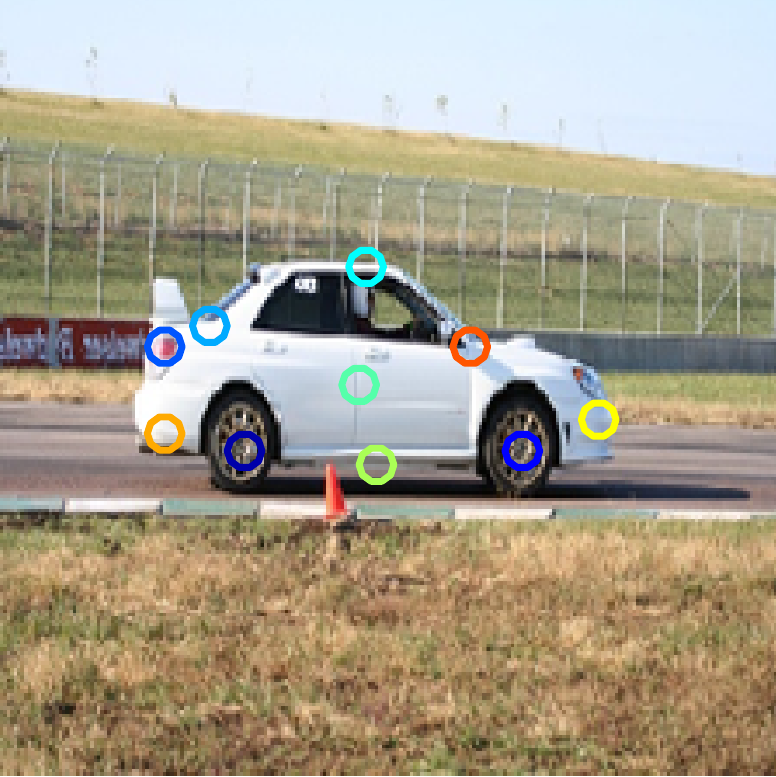} &
      \includegraphics[width=1\linewidth]{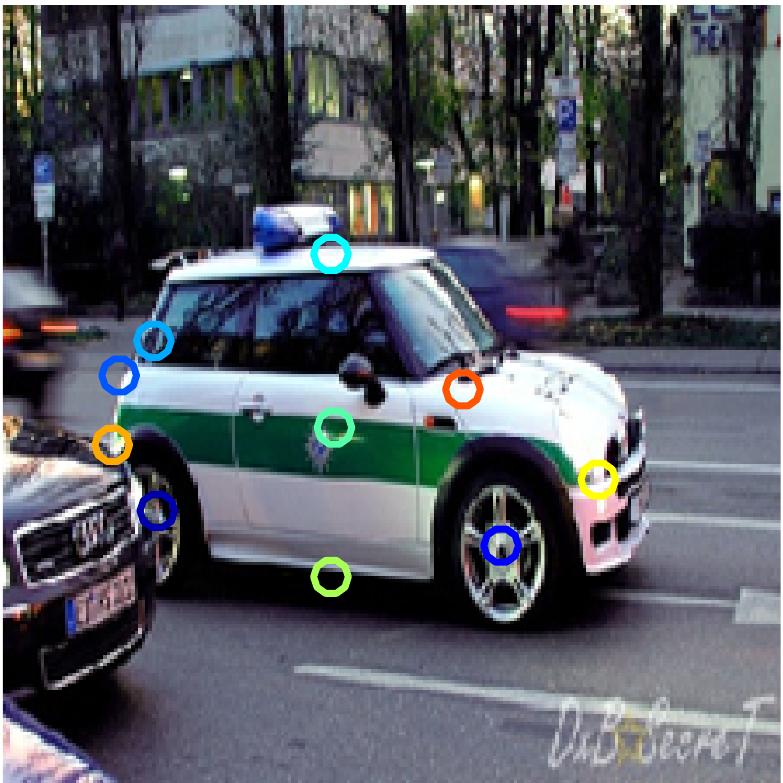} &
      \includegraphics[width=1\linewidth]{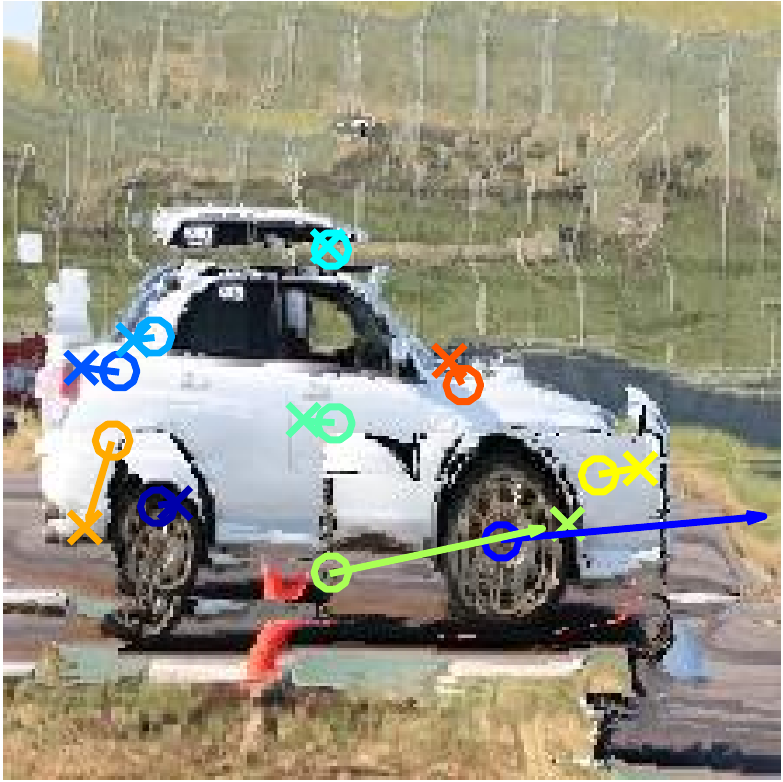}  &
      \includegraphics[width=1\linewidth]{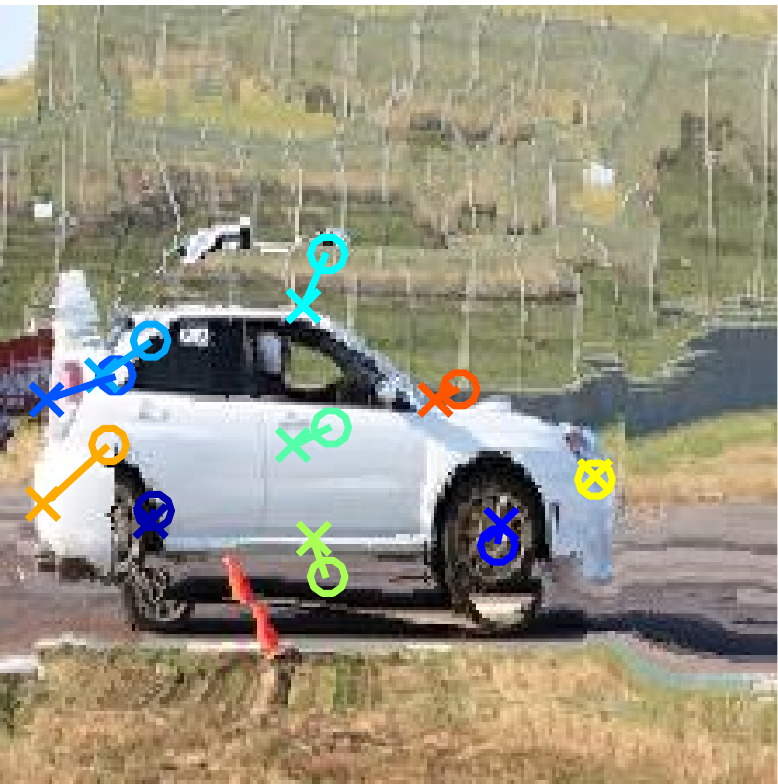} &
      \includegraphics[width=1\linewidth]{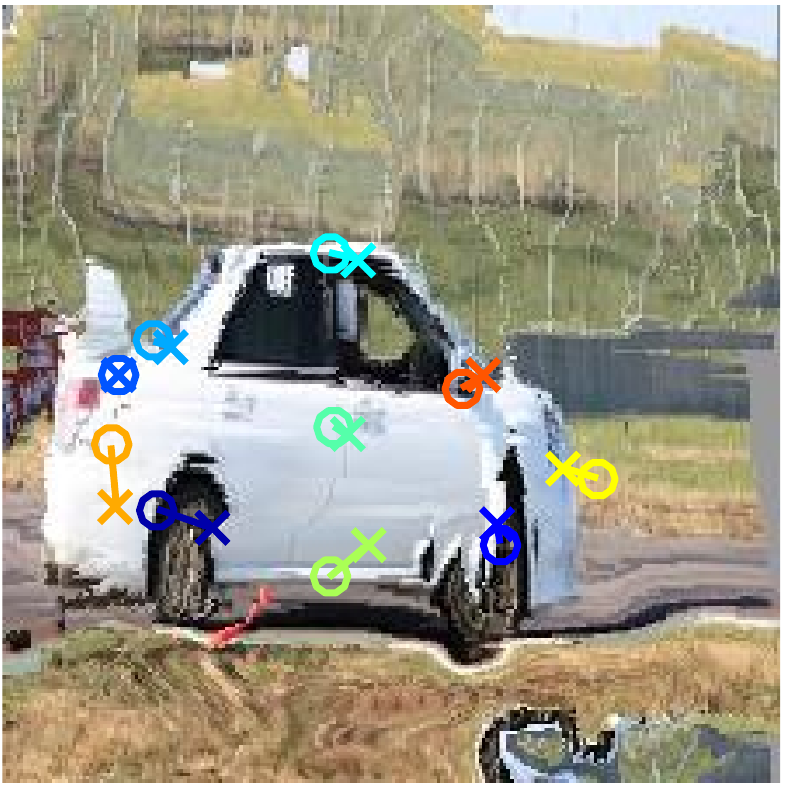} &
      \includegraphics[width=1\linewidth]{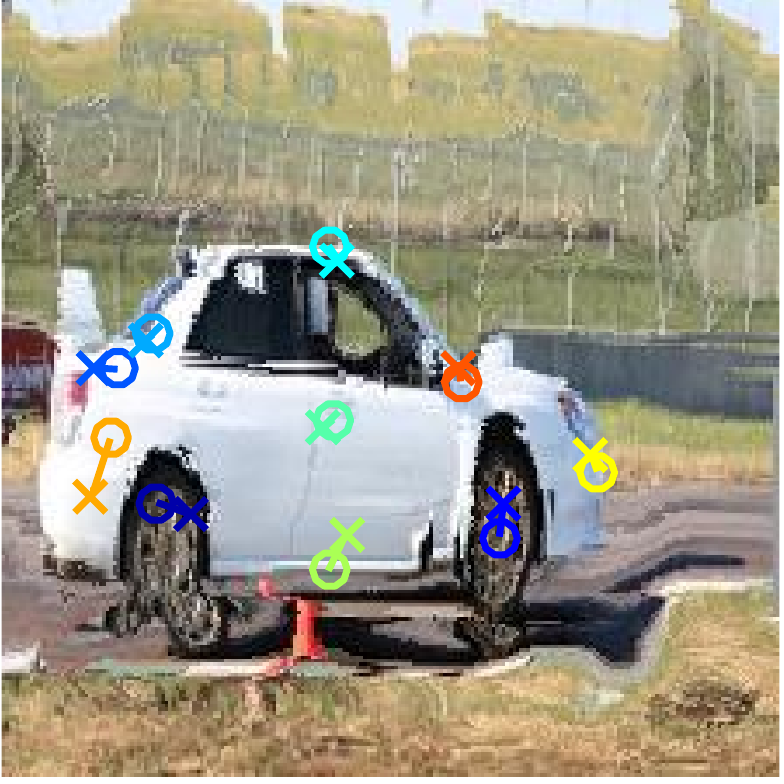} \\
      Source & Target & NAM$_{\rm HOG}$ & LOM$_{\rm HOG}$ & SCNet-A & SCNet-AG+ \\
   \end{tabular}                      
   \caption{Quantitative comparison of dense correspondence. We show the keypoints of the target image in circles and the predicted keypoints of the source in crosses, with a vector that depicts the matching error. (Best viewed in pdf.)}
 \label{fig:dense_match}
\end{figure*}

\vspace{-0.3cm}
\paragraph{Qualitative comparison.}
Region matching results for NAM, SCNet-A, and SCNet-AG+ are shown in Figure~\ref{fig:qanli_PF_PASCAL}. In this example, at the IoU threshold $0.5$, the numbers of correct matches are shown for all methods. We can see that SCNet models perform significantly better than NAM with HOG feature, and SCNet-A is outperformed by SCNet-AG+ that learns a geometric consistency term. 

\subsection{Flow field}
Given a sparse region matching result and its corresponding scores, we generate dense semantic flow using a densifying technique in~\cite{HCSP16}. In brief, we select out a region match with the highest score, and assign dense correspondences to the pixels within the matched regions by linear interpolation. This process is repeated without replacement of the region match until we assign dense correspondences to all pixels in the source image. The results are evaluated on PF-PASCAL dataset. To evaluate transferability performance of the models, we also test them on other datasets such as PF-WILLOW~\cite{HCSP16}, Caltech-101~\cite{fei2006one} and PASCAL Parts~\cite{zhou2015flowweb} datasets, 
and compare with state-of-the-art results on these datasets. In these cases direct comparison between learning-based methods may not be fair in the sense that they are trained on different datasets.

\begin{table*}[!htb]
\tabcolsep=0.001cm
\setlength{\tabcolsep}{.15em}
\small
\begin{minipage}{.4\linewidth}
\centering
\caption{Fixed-threshold PCK on PF-WILLOW.}
\label{pckWillow:table}

\begin{tabular}{l|ccc}
  \hline
  Method & PCK@0.05 & PCK@0.1 & PCK@0.15\\
  \hline
  SIFT Flow~\cite{liu2011sift}       & 0.247 & 0.380 & 0.504 \\
  DAISY w/SF~\cite{yang2014daisy}      & 0.324 & 0.456 & 0.555 \\
  DeepC w/SF~\cite{Zagoruyko15CVPR}   & 0.212 & 0.364 & 0.518 \\
  LIFT w/SF~\cite{2016lift}   & 0.224 & 0.346 & 0.489 \\
  VGG w/SF~\cite{SiZi2014vgg}        & 0.224 & 0.388 & 0.555 \\
  FCSS w/SF~\cite{kim2017fcss}       & 0.354 & 0.532 & 0.681 \\
  FCSS w/PF~\cite{kim2017fcss}       & 0.295 & 0.584 & 0.715 \\
  LOM$_{\rm HOG}$~\cite{HCSP16}       & 0.284   & 0.568 & 0.682    \\
  UCN\cite{UCN16}       & 0.291 & 0.417 & 0.513 \\
  SCNet-A         & 0.390 & {\bf 0.725} & {\bf 0.873} \\
  SCNet-AG          & {\bf 0.394} & 0.721 & 0.871 \\
  SCNet-AG+         & 0.386 & 0.704 & 0.853 \\
  \hline 
\end{tabular}
\end{minipage}\hfill
\begin{minipage}{.3\linewidth}
\centering

\caption{Results on Caltech-101.}
\label{clatech-101}

\begin{tabular}{l|ccc}
  \hline
  Methods   & LT-ACC & IoU  & LOC-ERR \\ \hline
  NAM$_{\rm HOG}$~\cite{HCSP16}       & 0.70   & 0.44 & 0.39    \\ 
  PHM$_{\rm HOG}$~\cite{HCSP16}       & 0.75   & 0.48 & 0.31    \\ 
  LOM$_{\rm HOG}$~\cite{HCSP16}       & 0.78   & 0.50 & 0.26    \\ 
  DeepFlow~\cite{weinzaepfel2015deepmatching}  &0.74& 0.40& 0.34\\ 
  SIFT Flow~\cite{liu2011sift}        & 0.75 & 0.48 & 0.32\\ 
  DSP~\cite{kim2013deformable}        & 0.77 & 0.47 & 0.35\\ 
  FCSS w/SF~\cite{kim2017fcss}        & 0.80   & 0.50  &{\bf0.21} \\
  FCSS w/PF~\cite{kim2017fcss}        & {\bf0.83}   &{\bf0.52} &0.22 \\
  SCNet-A   & 0.78   & 0.50 & 0.28    \\ 
  SCNet-AG  & 0.78   & 0.50 & 0.27    \\ 
  SCNet-AG+ &  0.79   & 0.51 & 0.25   \\ \hline
\end{tabular}
\end{minipage}\hfill
\begin{minipage}{.29\linewidth}
\centering

\caption{Results on PASCAL Parts.}
\label{pascal-parts}

  \begin{tabular}{l|ccc}
  \hline
  Methods   & IoU  & PCK \\ \hline
  NAM$_{\rm HOG}$~\cite{HCSP16}       & 0.35   & 0.13 \\
  PHM$_{\rm HOG}$~\cite{HCSP16}       & 0.39   & 0.17 \\
  LOM$_{\rm HOG}$~\cite{HCSP16}       & 0.41   & 0.17 \\
  Congealing~\cite{learned2006data}   & 0.38   & 0.11 \\
  RASL~\cite{peng2012rasl}            & 0.39   & 0.16 \\
  CollectionFlow~\cite{kemelmacher2012collection}  & 0.38  & 0.12  \\
  DSP~\cite{kim2013deformable}        & 0.39   & 0.17 \\
  FCSS w/SF~\cite{kim2017fcss}        & 0.44   & 0.28 \\
  FCSS w/PF~\cite{kim2017fcss}        & 0.46   & {\bf 0.29} \\
  SCNet-A   & 0.47   & 0.17   \\
  SCNet-AG  & 0.47   & 0.17   \\ 
  SCNet-AG+ & {\bf 0.48}   & 0.18   \\ \hline
  \end{tabular}
\end{minipage}

\end{table*}

\vspace{-0.3cm}
\paragraph{Results on PF-PASCAL.}
We compare SCNet with Proposal Flow~\cite{HCSP16} and UCN \cite{UCN16} on the PF-PASCAL dataset, and summarize the result in Table \ref{pck:table}. The UCN is retrained using the code provided by the authors on the PF-PASCAL dataset for fair comparison. 
Using the raw network of \cite{UCN16} trained on a different subset of PASCAL yields as expected lower performance, with a mean PCK of 36.0 as opposed to the 55.6 obtained for the retrained network. The three variants of SCNet do consistently better than UCN as well as all methods in~\cite{HCSP16}, with a PCK of 66.3 or above.
Among all the methods, SCNet-AG+ performs best with a PCK of 72.2. Figure~\ref{fig:dense_match} presents two examples of dense matching for PF-PASCAL. Ground truth are presented as circles and predicted keypoints are presented as crosses. We observe a better performance of SCNet-AG and SCNet-AG+.

\vspace{-0.3cm}
\paragraph{Results on PF-WILLOW.}
For evaluating transferability, we test (PF-PASCAL trained) SCNet and UCN on the PF-WILLOW dataset~\cite{HCSP16} and compare the results with recent methods in Table~\ref{pckWillow:table} where PCK is averaged over all classes. The postfix `w/SF' and `w/PF' represent that matching is performed by SIFT Flow~\cite{liu2011sift} and Proposal Flow~\cite{HCSP16}, respectively. On this dataset where the data has a different distribution, SCNet-AG slightly outperforms the A and AG+ variants~(PCK$@0.05$). 
We observe that all SCNet models significantly outperform UCN, which is trained on the same dataset with the SCNet models, as well as other methods using hand-crafted features~\cite{liu2011sift,yang2014daisy,KiMiHaSo2015dasc} and learned features~\cite{simo2015discriminative,Zagoruyko15CVPR,han2015matchnet,2016lift,SiZi2014vgg,kim2017fcss}. 

\vspace{-0.3cm}
\paragraph{Results on Caltech-101.}
We also evaluate our approach on the Caltech-101
dataset~\cite{fei2006one}. Following the experimental protocol
in~\cite{kim2013deformable}, we randomly select 15 pairs of images for
each object class, and evaluate matching accuracy with three metrics:~Label transfer accuracy (LT-ACC)~\cite{liu2011nonparametric}, the IoU
metric, and the localization error (LOC-ERR) of corresponding pixel
positions. Table~\ref{clatech-101} shows that SCNet achieves comparable results with the state of the art. 
The best performer, FCSS~\cite{kim2017fcss}, is trained on images from the same Caltech-101 dataset, while SCNet models are not. 


\vspace{-0.3cm}
\paragraph{Results on PASCAL Parts.}
Following \cite{HCSP16}, we use the dataset provided by \cite{zhou2015flowweb} where the images are sampled from the PASCAL part dataset \cite{chen2014detect}. For this experiment, we measure the weighted IoU score between transferred segments and the ground truth, with weights determined by the pixel area of each part. To evaluate alignment accuracy, we measure the PCK metric ($\alpha = 0.05$) using keypoint annotations for the PASCAL classes. Following \cite{HCSP16} once again,  we use selective search (SS) to generate proposals for SCNet in this experiment. The results are summarized in Table~\ref{pascal-parts}. SCNet models outperform all other results on the dataset in IoU, and SCNet-AG+ performs best among them. FCSS w/PF~\cite{kim2017fcss} performs better in PCK on this dataset. 

These results verify that SCNet models have successfully learned semantic correspondence. 



\section{Conclusion}
We have introduced a novel model for learning semantic correspondence, and proposed the corresponding CNN architecture that uses object proposals as matching primitives and learns matching in terms of appearance and geometry. The proposed method substantially outperforms both recent deep learning architectures and previous methods based on hand-crafted features. The result clearly demonstrates the effectiveness of learning geometric matching for semantic correspondence. In future work, we will explore better models and architectures to leverage geometric information.

\paragraph{Acknowledgments.}
This work was supported by the ERC grants VideoWorld and Allegro, the Institut Universitaire de France, the National Research Foundation of Korea (NRF) grant funded by the Korea government (MSIP) (No. 2017R1C1B2005584) as well as the MSIT (Ministry of Science and ICT), Korea, under the ICT Consilience Creative program (IITP-2017-R0346-16-1007). We gratefully acknowledge the support of NVIDIA Corporation with the donation of a Titan X Pascal GPU used for this research. We also thank JunYoung Gwak and Christopher B. Choy for their help in comparing with UCN.

{\small
\bibliographystyle{ieee}
\bibliography{match}

\begin{thebibliography}{10}\itemsep=-1pt

\bibitem{arbelaez2014multiscale}
P.~Arbelaez, J.~Pont-Tuset, J.~Barron, F.~Marques, and J.~Malik.
\newblock Multiscale combinatorial grouping.
\newblock In {\em Proc. IEEE Conf. Comp. Vision Patt. Recog.}, 2014.

\bibitem{BristowVL15}
H.~Bristow, J.~Valmadre, and S.~Lucey.
\newblock Dense semantic correspondence where every pixel is a classifier.
\newblock In {\em Proc. Int. Conf. Comp. Vision}, 2015.

\bibitem{chen2014detect}
X.~Chen, R.~Mottaghi, X.~Liu, S.~Fidler, R.~Urtasun, et~al.
\newblock Detect what you can: Detecting and representing objects using
  holistic models and body parts.
\newblock In {\em Proc. IEEE Conf. Comp. Vision Patt. Recog.}, 2014.

\bibitem{CKSP15}
M.~Cho, S.~Kwak, C.~Schmid, and J.~Ponce.
\newblock Unsupervised object discovery and localization in the wild:
  Part-based matching with bottom-up region proposals.
\newblock In {\em Proc. IEEE Conf. Comp. Vision Patt. Recog.}, 2015.

\bibitem{UCN16}
C.~Choy, J.~Gwak, S.~Savarese, and M.~Chandraker.
\newblock Universal correspondence network.
\newblock In {\em Proc. Neural Info. Proc. Systems}, 2016.

\bibitem{dalal2005histograms}
N.~Dalal and B.~Triggs.
\newblock Histograms of oriented gradients for human detection.
\newblock In {\em Proc. IEEE Conf. Comp. Vision Patt. Recog.}, 2005.

\bibitem{everingham2008pascal}
M.~Everingham, L.~Van~Gool, C.~Williams, J.~Winn, and A.~Zisserman.
\newblock The pascal visual object classes challenge 2007 (voc2007) results.

\bibitem{fei2006one}
L.~Fei-Fei, R.~Fergus, and P.~Perona.
\newblock One-shot learning of object categories.
\newblock {\em IEEE Trans. Patt. Anal. Mach. Intell.}, 28(4):594--611, 2006.

\bibitem{fischer2015flownet}
P.~Fischer, A.~Dosovitskiy, E.~Ilg, P.~H{\"a}usser, C.~Haz{\i}rba{\c{s}},
  V.~Golkov, P.~van~der Smagt, D.~Cremers, and T.~Brox.
\newblock Flownet: Learning optical flow with convolutional networks.
\newblock In {\em Proc. IEEE Conf. Comp. Vision Patt. Recog.}, 2015.

\bibitem{girshickICCV15fastrcnn}
R.~Girshick.
\newblock Fast r-cnn.
\newblock In {\em Proc. Int. Conf. Comp. Vision}, 2015.

\bibitem{HCSP16}
B.~Ham, M.~Cho, C.~Schmid, and J.~Ponce.
\newblock Proposal flow.
\newblock In {\em Proc. IEEE Conf. Comp. Vision Patt. Recog.}, 2016.

\bibitem{han2015matchnet}
X.~Han, T.~Leung, Y.~Jia, R.~Sukthankar, and A.~C. Berg.
\newblock Match{N}et: Unifying feature and metric learning for patch-based
  matching.
\newblock In {\em Proc. IEEE Conf. Comp. Vision Patt. Recog.}, 2015.

\bibitem{hariharan2012discriminative}
B.~Hariharan, J.~Malik, and D.~Ramanan.
\newblock Discriminative decorrelation for clustering and classification.
\newblock In {\em Proc. European Conf. Comp. Vision}, pages 459--472. Springer,
  2012.

\bibitem{kaiming14ECCV}
K.~He, X.~Zhang, S.~Ren, and J.~Sun.
\newblock Spatial pyramid pooling in deep convolutional networks for visual
  recognition.
\newblock In {\em Proc. European Conf. Comp. Vision}, 2014.

\bibitem{horn1993determining}
B.~K. Horn and B.~G. Schunck.
\newblock Determining optical flow: A retrospective.
\newblock {\em Artificial Intelligence}, 59(1):81--87, 1993.

\bibitem{hosang2015what}
J.~Hosang, R.~Benenson, P.~Doll\'ar, and B.~Schiele.
\newblock What makes for effective detection proposals?
\newblock {\em IEEE Trans. Patt. Anal. Mach. Intell.}, 2015.

\bibitem{hur2015generalized}
J.~Hur, H.~Lim, C.~Park, and S.~C. Ahn.
\newblock Generalized deformable spatial pyramid: Geometry-preserving dense
  correspondence estimation.
\newblock In {\em Proc. IEEE Conf. Comp. Vision Patt. Recog.}, 2015.

\bibitem{kanazawa2016warpnet}
A.~Kanazawa, D.~W. Jacobs, and M.~Chandraker.
\newblock Warp{N}et: Weakly supervised matching for single-view reconstruction.
\newblock In {\em Proc. IEEE Conf. Comp. Vision Patt. Recog.}, 2016.

\bibitem{kemelmacher2012collection}
I.~Kemelmacher-Shlizerman and S.~M. Seitz.
\newblock Collection flow.
\newblock In {\em Proc. IEEE Conf. Comp. Vision Patt. Recog.}, 2012.

\bibitem{kim2013deformable}
J.~Kim, C.~Liu, F.~Sha, and K.~Grauman.
\newblock Deformable spatial pyramid matching for fast dense correspondences.
\newblock In {\em Proc. IEEE Conf. Comp. Vision Patt. Recog.}, 2013.

\bibitem{kim2017fcss}
S.~Kim, D.~Min, B.~Ham, S.~Jeon, S.~Lin, and K.~Sohn.
\newblock Fcss: Fully convolutional self-similarity for dense semantic
  correspondence.
\newblock In {\em Proc. IEEE Conf. Comp. Vision Patt. Recog.}, 2017.

\bibitem{KiMiHaSo2015dasc}
S.~W. Kim, D.~Min, B.~Ham, and K.~Sohn.
\newblock Dasc: Dense adaptative self-correlation descriptor for multi-modal
  and multi-spectral correspondence.
\newblock In {\em Proc. IEEE Conf. Comp. Vision Patt. Recog.}, 2015.

\bibitem{krizhevsky2012imagenet}
A.~Krizhevsky, I.~Sutskever, and G.~E. Hinton.
\newblock Imagenet classification with deep convolutional neural networks.
\newblock In {\em Proc. Neural Info. Proc. Systems}, 2012.

\bibitem{learned2006data}
E.~G. Learned-Miller.
\newblock Data driven image models through continuous joint alignment.
\newblock {\em IEEE Trans. Patt. Anal. Mach. Intell.}, 28(2):236--250, 2006.

\bibitem{liu2011nonparametric}
C.~Liu, J.~Yuen, and A.~Torralba.
\newblock Nonparametric scene parsing via label transfer.
\newblock {\em IEEE Trans. Patt. Anal. Mach. Intell.}, 33(12):2368--2382, 2011.

\bibitem{liu2011sift}
C.~Liu, J.~Yuen, and A.~Torralba.
\newblock {SIFT} flow: Dense correspondence across scenes and its applications.
\newblock {\em IEEE Trans. Patt. Anal. Mach. Intell.}, 33(5):978--994, 2011.

\bibitem{long2014convnets}
J.~L. Long, N.~Zhang, and T.~Darrell.
\newblock Do convnets learn correspondence?
\newblock In {\em Proc. Neural Info. Proc. Systems}, 2014.

\bibitem{lowe2004distinctive}
D.~G. Lowe.
\newblock Distinctive image features from scale-invariant keypoints.
\newblock {\em Int. J. of Comp. Vision}, 60(2):91--110, 2004.

\bibitem{manen2013prime}
S.~Manen, M.~Guillaumin, and L.~Van~Gool.
\newblock Prime object proposals with randomized {P}rim's algorithm.
\newblock In {\em Proc. Int. Conf. Comp. Vision}, 2013.

\bibitem{matas2004robust}
J.~Matas, O.~Chum, M.~Urban, and T.~Pajdla.
\newblock Robust wide-baseline stereo from maximally stable extremal regions.
\newblock {\em Image and vision computing}, 22(10):761--767, 2004.

\bibitem{okutomi1993multiple}
M.~Okutomi and T.~Kanade.
\newblock A multiple-baseline stereo.
\newblock {\em IEEE Trans. Patt. Anal. Mach. Intell.}, 15(4):353--363, 1993.

\bibitem{peng2012rasl}
Y.~Peng, A.~Ganesh, J.~Wright, W.~Xu, and Y.~Ma.
\newblock Rasl: Robust alignment by sparse and low-rank decomposition for
  linearly correlated images.
\newblock {\em IEEE Trans. Patt. Anal. Mach. Intell.}, 34(11):2233--2246, 2012.

\bibitem{weinzaepfel2015deepmatching}
J.~{Revaud}, P.~{Weinzaepfel}, Z.~{Harchaoui}, and C.~{Schmid}.
\newblock Deepmatching: Hierarchical deformable dense matching.
\newblock {\em ArXiv e-prints}, 2015.

\bibitem{rhemann2011fast}
C.~Rhemann, A.~Hosni, M.~Bleyer, C.~Rother, and M.~Gelautz.
\newblock Fast cost-volume filtering for visual correspondence and beyond.
\newblock In {\em Proc. IEEE Conf. Comp. Vision Patt. Recog.}, 2011.

\bibitem{simo2015discriminative}
E.~Simo-Serra, E.~Trulls, L.~Ferraz, I.~Kokkinos, P.~Fua, and F.~Moreno-Noguer.
\newblock Discriminative learning of deep convolutional feature point
  descriptors.
\newblock In {\em Proc. Int. Conf. Comp. Vision}, 2015.

\bibitem{SiZi2014vgg}
K.~Simonyan and andrew Zisserman.
\newblock Very deep convolutional networks for large-scale visual recognition.
\newblock In {\em Proc. IEEE Conf. Comp. Vision Patt. Recog.}, 2014.

\bibitem{taniai2016joint}
T.~Taniai, S.~N. Sinha, and Y.~Sato.
\newblock Joint recovery of dense correspondence and cosegmentation in two
  images.
\newblock In {\em Proc. IEEE Conf. Comp. Vision Patt. Recog.}, 2016.

\bibitem{tau2014dense}
M.~Tau and T.~Hassner.
\newblock Dense correspondences across scenes and scales.
\newblock {\em IEEE Trans. Patt. Anal. Mach. Intell.}, 2015.

\bibitem{tola2010daisy}
E.~Tola, V.~Lepetit, and P.~Fua.
\newblock Daisy: An efficient dense descriptor applied to wide-baseline stereo.
\newblock {\em IEEE Trans. Patt. Anal. Mach. Intell.}, 32(5):815--830, 2010.

\bibitem{uijlings2013selective}
J.~R. Uijlings, K.~E. van~de Sande, T.~Gevers, and A.~W. Smeulders.
\newblock Selective search for object recognition.
\newblock {\em Int. J. of Comp. Vision}, 104(2):154--171, 2013.

\bibitem{Uijlings13ijcv}
J.~R.~R. Uijlings, K.~E.~A. van~de Sande, T.~Gevers, and A.~W.~M. Smeulders.
\newblock Selective search for object recognition.
\newblock {\em Int. J. of Comp. Vision}, 104(2):154--171, 2013.

\bibitem{weinzaepfel2013deepflow}
P.~Weinzaepfel, J.~Revaud, Z.~Harchaoui, and C.~Schmid.
\newblock Deepflow: Large displacement optical flow with deep matching.
\newblock In {\em Proc. Int. Conf. Comp. Vision}, 2013.

\bibitem{yang2014daisy}
H.~Yang, W.-Y. Lin, and J.~Lu.
\newblock Daisy filter flow: A generalized discrete approach to dense
  correspondences.
\newblock In {\em Proc. IEEE Conf. Comp. Vision Patt. Recog.}, 2014.

\bibitem{yang2013articulated}
Y.~Yang and D.~Ramanan.
\newblock Articulated human detection with flexible mixtures of parts.
\newblock {\em IEEE Trans. Patt. Anal. Mach. Intell.}, 35(12):2878--2890, 2013.

\bibitem{2016lift}
K.~M. Yi, E.~Trulls, V.~Lepetit, and P.~Fua.
\newblock Lift: Learned invariant feature transform.
\newblock In {\em Proc. European Conf. Comp. Vision}, 2016.

\bibitem{Zagoruyko15CVPR}
S.~Zagoruyko and N.~Komodakis.
\newblock Learning to compare image patches via convolutional neural networks.
\newblock In {\em Proc. IEEE Conf. Comp. Vision Patt. Recog.}, 2015.

\bibitem{vzbontar2014computing}
J.~{\v{Z}}bontar and Y.~LeCun.
\newblock Computing the stereo matching cost with a convolutional neural
  network.
\newblock In {\em Proc. IEEE Conf. Comp. Vision Patt. Recog.}, 2015.

\bibitem{zbontar2016stereo}
J.~Zbontar and Y.~LeCun.
\newblock Stereo matching by training a convolutional neural network to compare
  image patches.
\newblock {\em Journal of Machine Learning Research}, 17(1-32):2, 2016.

\bibitem{zhou2015flowweb}
T.~Zhou, Y.~Jae~Lee, S.~X. Yu, and A.~A. Efros.
\newblock Flow{W}eb: Joint image set alignment by weaving consistent,
  pixel-wise correspondences.
\newblock In {\em Proc. IEEE Conf. Comp. Vision Patt. Recog.}, 2015.

\bibitem{zhou2016learning}
T.~Zhou, P.~Kr{\"a}henb{\"u}hl, M.~Aubry, Q.~Huang, and A.~A. Efros.
\newblock Learning dense correspondence via 3d-guided cycle consistency.
\newblock In {\em Proc. IEEE Conf. Comp. Vision Patt. Recog.}, 2016.

\bibitem{zitnick2014edge}
C.~L. Zitnick and P.~Doll{\'a}r.
\newblock Edge boxes: Locating object proposals from edges.
\newblock In {\em Proc. European Conf. Comp. Vision}, 2014.

\end{thebibliography}
}

\end{document}